\documentclass[10pt,twocolumn,letterpaper]{article}
\usepackage{graphicx}
\usepackage{subcaption}
\usepackage{tikz}
\usepackage{float}
\usepackage{xr}
\usepackage{multirow}
\usepackage{tabularx} 

\usepackage{cvpr}              

\definecolor{cvprblue}{rgb}{0.21,0.49,0.74}
\usepackage[pagebackref,breaklinks,colorlinks,citecolor=cvprblue]{hyperref}
\usepackage[accsupp]{axessibility} 
\def\paperID{9394} 
\def\confName{CVPR}
\def\confYear{2024}

\title{DS-NeRV: Implicit Neural Video Representation with Decomposed Static and Dynamic Codes}
\author{Hao Yan \qquad Zhihui Ke \qquad Xiaobo Zhou$^{\dagger}$ \qquad Tie Qiu \qquad Xidong Shi \qquad Dadong Jiang\\
College of Intelligence and Computing, Tianjin University\\
{\tt\small {\{yan$\_$hao, kezhihui, xiaobo.zhou, qiutie, suif, patrickdd\}}@tju.edu.cn}
}
\begin{document}
\twocolumn[{
\maketitle
\begin{center}
\vspace{-9mm}
    \captionsetup{type=figure}
    \includegraphics[width=1.\textwidth]{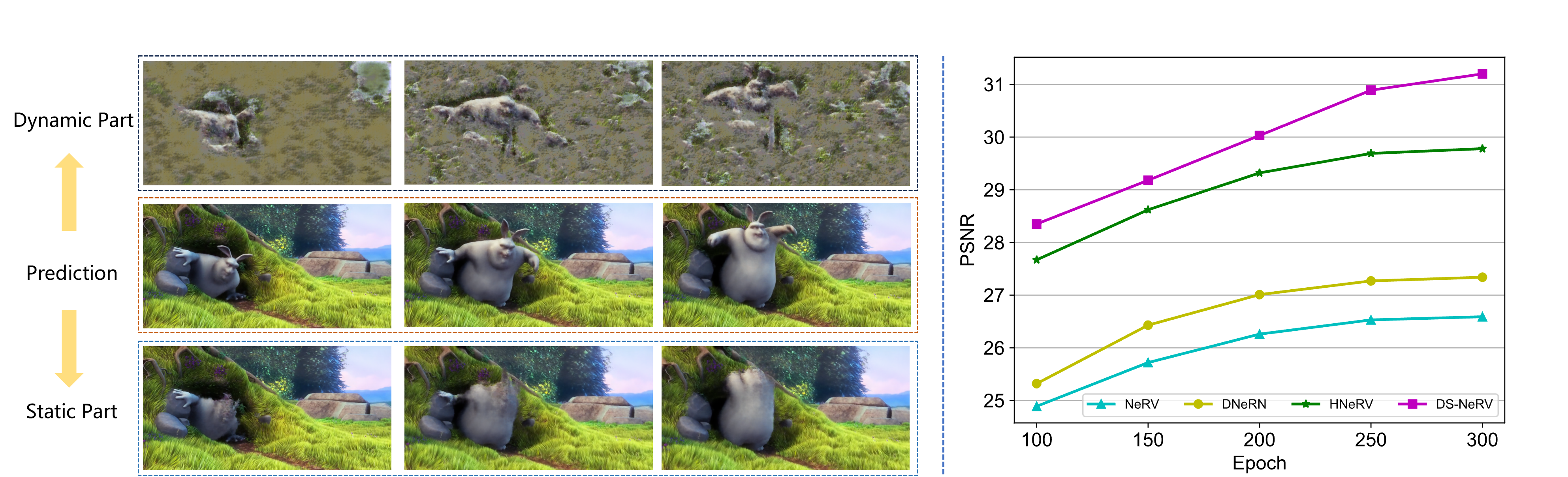}
    \captionof{figure}{(Left) Our proposed DS-NeRV decomposes the video into learnable static and dynamic codes, which represent static elements and dynamic elements in the video. (Right) Video reconstruction results for various implicit neural representations with 0.35M.}
    \label{fig:abstract}
\end{center}

}]
\begin{abstract}
\vspace{-4mm}

Implicit neural representations for video (NeRV) have recently become a novel way for high-quality video representation. 
However, existing works employ a single network to represent the entire video, which implicitly confuse static and dynamic information.
This leads to an inability to effectively compress the redundant static information and lack the explicitly modeling of global temporal-coherent dynamic details. 
To solve above problems, we propose DS-NeRV, which decomposes videos into sparse learnable static codes and dynamic codes without the need for explicit optical flow or residual supervision.
By setting different sampling rates for two codes and applying weighted sum and interpolation sampling methods, DS-NeRV efficiently utilizes redundant static information while maintaining high-frequency details.
Additionally, we design a cross-channel attention-based (CCA) fusion module to efficiently 
 fuse these two codes for frame decoding.
Our approach achieves a high quality reconstruction of 31.2 PSNR with only 0.35M parameters thanks to separate static and dynamic codes representation and outperforms existing NeRV methods in many downstream tasks. Our project website is at \href{https://haoyan14.github.io/DS-NeRV/}{https://haoyan14.github.io/DS-NeRV/}.
\end{abstract}
\let\thefootnote\relax\footnotetext{$\dagger$ Corresponding Author}

\externaldocument{main}

\section{Introduction}
In the first half of 2022, video traffic accounted for a substantial 65.93\% share of the overall network traffic and constituted as much as 80\% of the total downstream traffic during the evening peak hours~\cite{sandvine, cisco}. 
This causes tremendous pressure on network communication and storage. Thus, it is crucial to explore more efficient video representations for compression.



In recent years, implicit neural representations (INR) have emerged as a promising solution due to their remarkable capacity to represent diverse forms of signals~\cite{nerf, dnerf, nerv, siren}. 
With the development of INR, it has been applied to video representation tasks, such as NeRV~\cite{nerv}, which has transformed the challenge of video compression into a problem of model compression.
Additionally, INR-based video representations often exhibit a simpler training process and a higher decoding speed~\cite{hnerv} compared to traditional video compression methods~\cite{h264,hevc,mpeg,av1} and learning-based video compression methods~\cite{pcnn,hlvc,dvc,scale-space,m-lvc}.

Typically, INR-based video representations can be categorized into two types: (1) \textit{index-based}~\cite{nerv,enerv,psnerv} methods that model the video as a neural network, where the positional encoding of the frame index is taken as input to reconstruct the corresponding frame.
(2) \textit{hybrid-based} methods~\cite{hnerv, dnerv} that employ an encoder-decoder architecture where they input each frame into the encoder to obtain the corresponding embedding, which is then forwarded to the decoder for reconstruction. Compared with index-based methods which is content-agnostic, hybrid-based methods leverage frame embedding to encapsulate frame information, thereby enhancing reconstruction quality.
However, the aforementioned two methods model the video as a whole, confusing both static and dynamic information within the video implicitly in the model parameters. 
Therefore, they cannot effectively compress static redundant information and model globally  coherent dynamic elements in the video.

Typically, a video consists of time-invariant static elements and time-varying dynamic elements. As shown in Fig.~\ref{fig:abstract} (Left), the grassy and rocks in the background either remain static or change minimally, while the bunny's posture exhibits noticeable changes over time.
Thus, to reduce the size of the video INR, it is beneficial to compress these redundant static information. On the other hand, the dynamic elements require smooth modeling across the entire video to preserve high-frequency details. 
In this paper, we draw inspiration from the above insight and propose DS-NeRV, a method that decomposes video into sparse learnable static codes $\mathcal{C}^s$ and dynamic codes $\mathcal{C}^d$, which respectively represent the static and dynamic elements in the video. The design of the learnable codes bears resemblance to the learnable noise vector used in \textit{Generative Latent Optimization} GLO)~\cite{glo}.
By assigning different sampling rates and sampling methods for two codes, 
DS-NeRV effectively decomposes the static and dynamic components of the video without the need for explicit optical flow or residual supervision and compresses redundant static information while preserving high-frequency dynamic details.
For a given frame index $t$, we compute the corresponding static code by finding the two closest static codes $c^s_i$ and $c^s_j$, then performing a weighted sum based on their distances. The corresponding dynamic code is obtained by interpolating dynamic codes $\mathcal{C}^d$ to the video's length and then selecting the corresponding code with index $t$.
In addition, we propose a cross-channel attention-based (CCA) fusion mechanism for efficiently fusing static and dynamic codes.

In summary, our contributions are as follows:
\begin{itemize}
    \item We propose DS-NeRV, a novel video INR, that decomposes the video into sparse learnable static and dynamic codes, which respectively represent static and dynamic elements within the video. This decomposition appears without the need for explicit optical flow or residual supervision.
    
    \item We carefully design the different sampling rates and sampling strategies for two codes to efficiently exploit the characteristics of videos.
    Moreover, we develop a cross-channel attention-based fusion module to fuse static and dynamic codes for video decoding.
    \item We conduct extensive experiments on three datasets and various downstream tasks to validate the effectiveness of DS-NeRV. 
    The experimental results demonstrate that DS-NeRV achieves more efficient video modeling over existing INR methods through decomposed static and dynamic codes representation.
\end{itemize}
\begin{figure*}[tbp]
  \centering
    \includegraphics[width=1\textwidth]{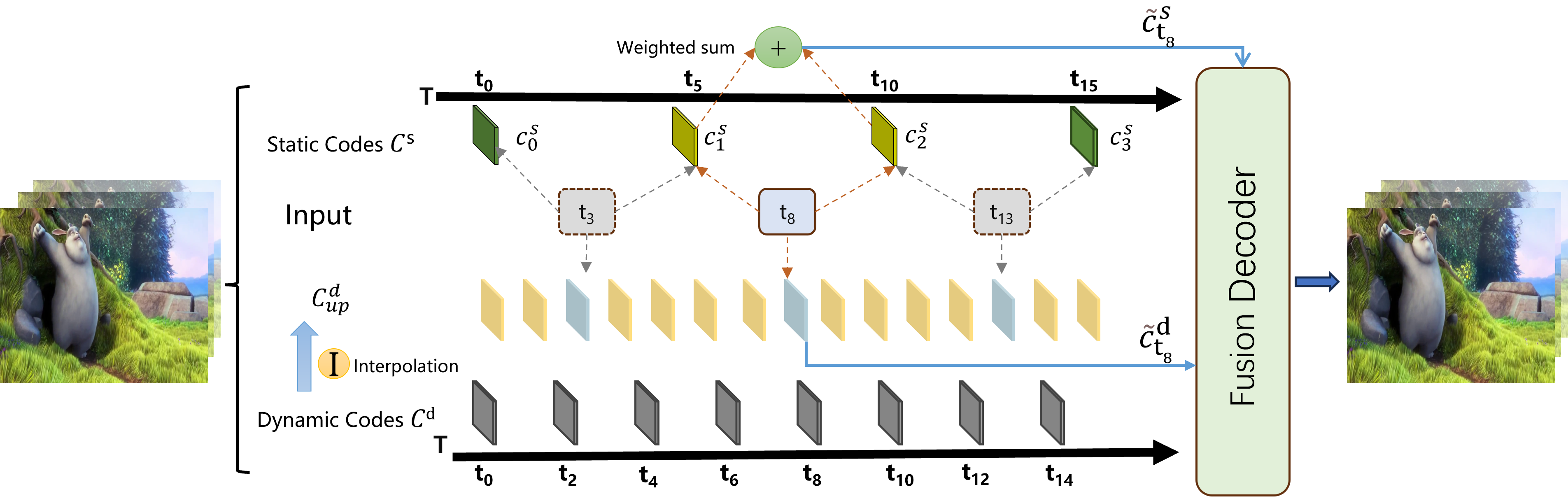}
    \caption{\textbf{DS-NeRV framework overview}. DS-NeRV decomposes the video into learnable static and dynamic codes.
    \textbf{Static Codes.} The two \textcolor[RGB]{205,205,0}{orange} static codes shown above are the two nearest selected. After weighted sum, they are forwarded to the fusion decoder. \textbf{Dynamic Codes.} We interpolate the dynamic codes to match the length of the video. Then the dynamic code corresponding to $t$ is selected in \textcolor[RGB]{102,159,180}{blue}.
    }
    \label{fig:overviewandselect}
    \vspace{-4mm}
\end{figure*}

\section{Related Work}
\label{sec:related}
\textbf{Implicit Neural Representation.} 
The purpose of INR is to model various signals through a function $\mathcal{F}$ that maps the input coordinate $\theta$ to corresponding value $y=\mathcal{F}(\theta),\theta \in \mathbb{R}^n, y \in \mathbb{R}^m$.
Starting from NeRF~\cite{nerf}, INR combined with neural rendering methods have developed rapidly in the field of novel view synthesis for static~\cite{mipnerf,ingp,few-view,tensorf,nope} and dynamic~\cite{dnerf,nrnerf,nerfflow,kplanes} scenes, and 3D reconstruction~\cite{deepsdf,occupancy}.
Recently, INR have been increasingly applied in the video representation. 
Different from Siren~\cite{siren} which maps frame pixel coordinates to their corresponding RGB, NeRV~\cite{nerv} introduces an approach by mapping frame index directly to corresponding video frame, thus enhancing both efficiency and performance. The proposal of NeRV promoted the development of INR for video~\cite{cnerv, enerv, psnerv, hnerv, d-nerv, dnerv, nirvana, nvp, ffnerv}. 
In contrast to existing studies that model the video as a whole, DS-NeRV decomposes the video into learnable static and dynamic codes, both of which are jointly learned during training. Thus, DS-NeRV can be seen as a novel INR for videos.

\noindent\textbf{Video Compression.} Traditional video compression methods (\eg~H.264~\cite{h264}, HEVC~\cite{hevc}) utilize predictive coding architectures to encode motion information and residual data of videos.
With the development of deep learning, video compression algorithms based on neural networks~\cite{hlvc,dvc,cu,entropycoding,elf,deepcoder,dcvc,wu_eccv} have garnered significant attention. However, these methods are limited to the conventional video compression workflow, severely impacting their capabilities. In NeRV-like methods, the problem of video compression can be converted to a model compression problem. Through techniques such as model pruning, model quantization, and entropy encoding, DS-NeRV achieves comparable performance with traditional video compression approaches and other INR methods.

\noindent\textbf{Latent Optimization For Representation learning.} 
Latent Optimization is employed in generative adversarial networks (GAN) to enhance the quality of samples $z$~\cite{latent_optimisation}.
GLO~\cite{glo} constructs a learnable noise vector for each image in the dataset, thereby offering a novel approach for image generation.
This method has also been introduced in the field of novel view synthesis. 
To improve the reconstruction quality,~\cite{n3dv,nerf-w,blocknerf} parameterize scene motion and appearance changes with a compact set of latent codes.
Inspired from GLO, DS-NeRV models the static and dynamic elements of videos using learnable codes which resemble the learnable noise vector in GLO. In this way, DS-NeRV can achieve higher performance in an end-to-end training manner thanks to the greater expressive ability of the codes.

\section{Method}

\label{sec:method}

\subsection{Overview}\label{sec:overview}
Given a video sequence $\mathcal{V}=\{v_{t}\}_{t=0}^{T-1} \in \mathbb{R}^{T \times H \times W \times 3}$, our target is to reconstruct the frame $v_t$ based on the frame index $t$. To achieve this, we decompose the video into learnable static codes $\mathcal{C}^s \in \mathbb{R}^{l_{s} \times h_{s} \times w_{s} \times dim_{s}} $ and dynamic codes $\mathcal{C}^d \in \mathbb{R}^{l_{d} \times h_{d} \times w_{d} \times dim_{d}} $. Given the frame index $t$, we obtain the corresponding static code $\widetilde{c}^s_t$ by weighted sum and dynamic code $\widetilde{c}^d_t$ through interpolation. 
The obtained $\widetilde{c}^s_t$ and $\widetilde{c}^d_t$ are then forwarded to the fusion decoder module to reconstruct the frame $v_t$, as shown in Fig.~\ref{fig:overviewandselect}. 
\begin{figure*}[tbp]
  \centering
    \includegraphics[width=1\textwidth]{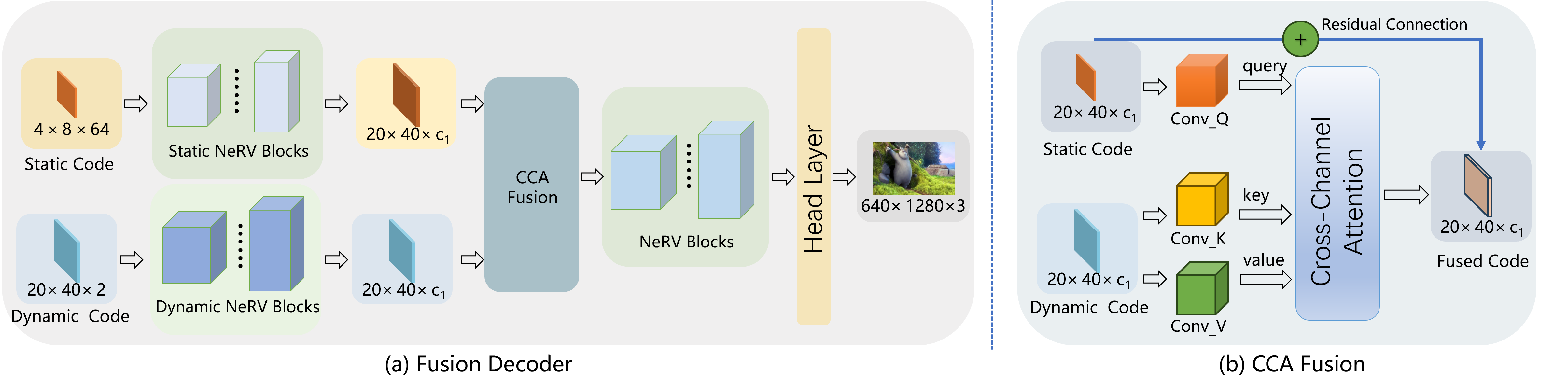}
    \caption{(a) The pipeline of Fusion Decoder. Decoder takes the static code and dynamic code corresponding to index $t$ as input and fuses their information to output frame. (b) Architecture of CCA Fusion Module. The CCA module fuses static code $\widetilde{c}^s_t$ and dynamic code $\widetilde{c}^d_t$ by cross-channel attention.}
  \label{fig:Decoder_architecture}
  \vspace{-4mm}
\end{figure*}
\subsection{Video Modeling}
\label{sec:modeling}
Traditional video compression pipelines~\cite{h264,hevc} use I-frames (Intra-frames) and P-frames (Predictive frames) for efficient video encoding and decoding.
The former contain complete information and are independent of other frames, serving as key reference points in the video sequence. On the other hand, the latter store motion and residual data, relying on the preceding decoded I-frames or P-frames for reference to decode.

Inspired by this design concept, we utilize static codes $\mathcal{C}^s$ with a low sampling rate $r_s$ to represent static elements in the video that can be shared to compress redundancy, while using dynamic codes $\mathcal{C}^d$ with a relatively high sampling rate $r_d$ to represent rich dynamic information. 

\noindent \textbf{Static Codes.} 
As \cref{fig:overviewandselect} (Top) shows, the static codes $\mathcal{C}^s = \{c^s_0, \cdots, c^s_i, \cdots, c^s_{l_s-1}\}$ is evenly distributed along the timeline at interval $z_s$. Consequently, given sampling rate $r_s$, the length of static codes is defined as $l_s = T \cdot r_s$, $l_s \ll T$ and the interval is computed as $z_s =  T / l_s $. More sampling details can be found in supplementary material.

According to E-NeRV~\cite{enerv}, the MLP used for feature map initialization before NeRV blocks often results in large parameters. To solve this problem, we prefer storing each static code $c^s_i$ in a 3D vector with dimensions $h_{s} \times w_{s} \times dim_{s}$, rather than a 1D vector which will be upsampled to initialize the feature map as adopted in~\cite{nerv,hnerv}. 
The 3D vector design eliminates the parameter overhead associated with the MLP before NeRV blocks. In our experiments, we set the size of each static code $c^s_i$ to $4\times8\times64$ for $960 \times 1920 \times 3$ video frame.

The frames between two adjacent static codes can be similarly considered as a GOP (Group of Pictures)~\cite{hevc} in HEVC, containing massive redundant static information that can be shared. So to effectively leverage the information stored in static codes, we design an innovative sampling method to obtain the static code $\widetilde{c}^s_t$ corresponding to frame index $t$. 
Given $t$, instead of solely relying on the nearest static code to obtain static information, we integrate information from two adjacent static codes, summing weighted by their respective distances to $t$.

As illustrated in Fig.~\ref{fig:overviewandselect}, for a given frame index $t$, we first obtain the two adjacent static codes indices $i$ and $j$ $(0\leq i<j< l_s)$ and then calculate their corresponding wights $w_i$ and $w_j$ according to their distances to $t$.
\begin{equation}
        dis_{i} = \mid t - ( i \cdot (z_s+1) ) \mid , dis_j = \mid t - ( j \cdot (z_s+1) ) \mid
  \label{eq:findnear}
\end{equation}
\begin{equation}
        w_{i} = \dfrac{dis_{j}}{(dis_{i} + dis_{j})},   
        w_{j} = \dfrac{dis_{i}}{(dis_{i} + dis_{j})}   
\end{equation}
Based on weights and indices, we then perform a weighted sum to obtain the final static code $\widetilde{c}^s_t$ as follows: 
\begin{equation}
  \widetilde{c}^s_t = w_{i} \cdot c^s_{i} + w_{j} \cdot c^s_{j}
  \label{eq:c_s_t}
\end{equation}
In this way, the associated static content can be computed for each frame index, effectively sharing static information throughout the video. Additionally, the sparse codes design also helps compress redundant static information.



\noindent\textbf{Dynamic Codes.} 
To characterize the rich dynamic information in the video, the length of dynamic codes is $l_d=T \cdot r_d$ with a higher sampling rate $r_d$. Similar to static codes representation, we store each dynamic code as a 3D vector to reduce the parameters. Therefore, we set the overall dynamic codes $\mathcal{C}^d$ with a size of $l_{d} \times h_{d} \times w_{d} \times dim_{d}$.
In our experiments, we set the size of each dynamic code to $20 \times40 \times 2$ for $960 \times 1920 \times 3$ video frame by default.

Different from the sampling method used in the static codes, we obtain the corresponding dynamic code $\widetilde{c}^d_t$ through the interpolated dynamic codes $\mathcal{C}^d_{up}$. The interpolation sampling method establishes global temporal coherence among dynamic codes through internal interaction, aligning with the perceptual continuity of motion in the real world. 

Specifically, we firstly interpolate the dynamic codes to match the length of the original video, while keeping the height, width, and number of channels unchanged. 
\begin{equation}
  \mathcal{C}^d_{up} = interpolate(\mathcal{C}^d,T,h_d,w_d,dim_d)
  \label{eq:c_dinterpolate}
\end{equation}
We subsequently retrieval the dynamic code $\widetilde{c}^d_t$ from the interpolated one $\mathcal{C}^d_{up}$ with index $t$, as depicted in Fig.~\ref{fig:overviewandselect}. 

Our dynamic codes representation offers low storage overhead by avoiding per-frame code storage while remaining compact total size and realizes the modeling of the global dynamic information through interpolation.
Moreover, the interpolation enables the generation of frames that were not seen during training, thereby supporting smooth and meaningful frame interpolation~\cite{nvp,glo}.
\subsection{Fusion Decoder}
\label{decoders}

\begin{figure*}
    \centering
    \includegraphics[width=0.99\textwidth]{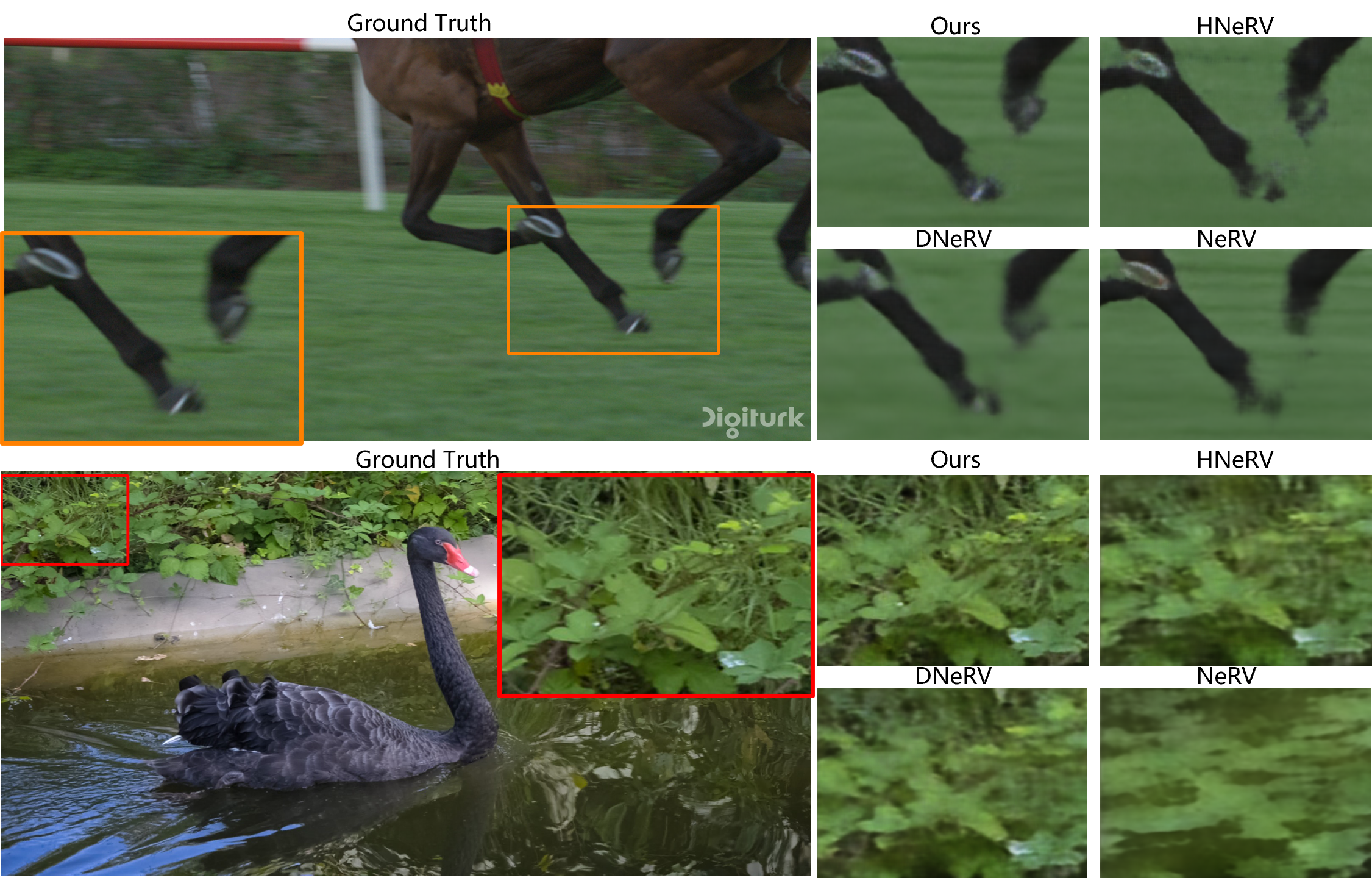}
     \caption{\textbf{Video reconstruction results on UVG and DAVIS.}  (Top) Jockey. (Bottom) Blackswan. 
     }
  \label{fig:uvg}
  \vspace{-3mm}
\end{figure*}

\textbf{Pipeline.} 
Since the obtained static code $\widetilde{c}^s_t$ and dynamic code $\widetilde{c}^d_t$ have different heights and widths, we firstly employ NeRV blocks to align their spatial dimensions, as shown in Fig.~\ref{fig:Decoder_architecture} (a). They are then forwarded to the cross-channel attention-based (CCA) module for fusion. Once the fusion module integrates information from $\widetilde{c}^s_t$ and $\widetilde{c}^d_t$, the fused code is then processed by the stacked NeRV Blocks to progressively upsample to the corresponding frame.

\noindent\textbf{CCA Fusion.} 
When considering fusion, a natural way to fuse $\widetilde{c}^s_t$ and $\widetilde{c}^d_t$ is to simply add them together.
However, this is not an appropriate approach~\cite{d-nerv} as they encode features from different domains, where static codes capture the static information but dynamic codes represent the motion-related information. To fuse the two types of information effectively, inspired by cross attention~\cite{cross} and channel attention~\cite{cbam}, we design a CCA fusion module based on cross-channel attention mechanism. 


Compared to the more commonly used spatial attention~\cite{vit}, we choose channel attention because during the CCA fusion stage, the spatial dimensions of $\widetilde{c}^s_t$ and $\widetilde{c}^d_t$ are identical, but their channels are different. 
We can think that at the same spatial position (u,v) in two codes, each code represents the static or dynamic information corresponding to the same region in original frame.
Therefore, in video representation task, we do not focus the interaction between different positions $(u_1, v_1)$ and $(u_2, v_2)$ in the two codes as this interaction does not contribute to the fusion between two features with the same spatial distribution.
Instead, we prioritize the interaction between different channels of the two codes given their distinct channel structures. Hence, we choose cross-channel attention to capture the information interaction between two codes for effective fusion.

Specifically, as illustrated in Fig.~\ref{fig:Decoder_architecture} (b), we treat each channel in the static code $\widetilde{c}^s_t$ as a query and each channel in the dynamic code $\widetilde{c}^d_t$ as a key-value pair. To achieve this, we firstly utilize three convolutions to extract the query, key, and value components from $\widetilde{c}^s_t$ and $\widetilde{c}^d_t$. 
Subsequently, we flatten these components along the spatial dimension to do channel attention, as follows:
\begin{equation}
  Q(t) = Flatten(Conv_q(\widetilde{c}^s_t))
  \label{eq:toq}
\end{equation}
\begin{equation}
  K(t),V(t) = Flatten(Conv_k(\widetilde{c}^d_t)),Flatten(Conv_v(\widetilde{c}^d_t))
  \label{eq:tok}
\end{equation}
After the attention mechanism, we integrate the static code $\widetilde{c}^s_t$ into the obtained attention output through residual connections.
\begin{equation}
  FusedCode(t) = softmax(QK^T)V + \widetilde{c}^s_t
  \label{eq:softmax}
\end{equation}

\section{Experiments}


\subsection{Setup}
\label{sec:setup}

\begin{table*}[!t]
    \begin{subtable}{.5\linewidth}
    \centering
    \tabcolsep=0.22cm
    \resizebox{!}{1.0cm}{
            \begin{tabular}{c|c c c c c } \hline 
            sizes& 0.35M& 0.75M& 1.5M&3M \\ \hline 
            NeRV\cite{nerv}& 26.59& 28.70& 30.60&34.37\\ 
            DNeRV\cite{dnerv}& 27.34& 30.01& 31.19&34.09\\
            HNeRV\cite{hnerv}& 29.78& 32.35& 35.20&37.74\\ \hline 
            \textbf{Ours}& \textbf{31.20}& \textbf{33.82}& \textbf{36.44}&\textbf{38.65}\\ \hline
            \end{tabular}
    }
    \caption{PSNR($\uparrow$) on Bunny with varying \textbf{model size}.}
    \label{tab:bunny}
    \end{subtable}%
    \begin{subtable}{.5\linewidth}
    \centering
    \tabcolsep=0.22cm
    \resizebox{!}{1.0cm}{
            \begin{tabular}{c|c c c c c}
            \hline
                epochs & 100 & 150 & 200 & 250 & 300  \\ \hline
                NeRV\cite{nerv} & 24.89 & 25.72 & 26.26 & 26.53 & 26.59  \\ 
                DNeRV\cite{dnerv} & 25.32 & 26.43 & 27.01 & 27.27 & 27.34  \\ 
                HNeRV\cite{hnerv} & 27.67 & 28.62 & 29.32 & 29.69 & 29.78  \\ \hline
                Ours & \textbf{28.35} & \textbf{29.18} & \textbf{30.03} & \textbf{30.89} & \textbf{31.20}  \\ \hline
            \end{tabular}
    }
    \caption{PSNR($\uparrow$) On Bunny with varying \textbf{epochs}.}
    \label{tab:epochs}
    \end{subtable}%

    \begin{subtable}{.5\linewidth}
    \centering
    \tabcolsep=0.05cm
    \resizebox{!}{1.0cm}{
            \begin{tabular}{c|c c c c c c c| c}
            \hline
            960x1920 & Beauty & Bosph & Honey & Jockey & Ready & Shake & Yacht & avg.\\ 
            \hline 
            NeRV\cite{nerv} & 33.33 & 33.34 & 38.79 & 28.97 & 23.89 & 33.89 & 27.05 & 31.32\\ 
            DNeRV\cite{dnerv} & 33.16 & 32.96 & 38.43 & 31.08 & 24.76 & 33.71 & 27.30 & 31.63\\
            HNeRV\cite{hnerv} & 33.88 & 35.02 & 39.41 & 31.69 & 25.72 & 34.95 & 29.09 & 32.82\\ 
            \hline 
            \textbf{Ours} & \textbf{33.97}& \textbf{35.22}& \textbf{39.56}& \textbf{32.86}& \textbf{27.10}& \textbf{35.04}& \textbf{29.40}& \textbf{33.31}\\
            \hline
            \end{tabular}
    }
    \caption{PSNR($\uparrow$) On UVG at resolution 960x1920.}
    \label{tab:uvg}
    \end{subtable}%
    \begin{subtable}{.5\linewidth}
    \tabcolsep=0.05cm
    \centering
    \resizebox{!}{1.0cm}{
            \begin{tabular}{c|c c c c c c c |c}
            \hline 
            480x960 & Beauty & Bosph & Honey & Jockey & Ready & Shake & Yacht & avg.\\ 
            \hline 
            NeRV\cite{nerv} & 34.98 & 34.98 & 40.73 & 31.23 & 24.92 & 34.95 & 28.59 & 32.91\\ 
            DNeRV\cite{dnerv} & 34.48 & 33.9 & 38.66 & 31.36 & 25.30 & 33.00 & 28.56 & 32.18\\
            HNeRV\cite{hnerv} & \textbf{35.42} & 36.13 & 41.47 & 32.64 & 26.54 & 36.04 & 30.22 & 34.07\\ 
            \hline 
            \textbf{Ours} & 35.37 & \textbf{36.25} & \textbf{41.67} & \textbf{33.48} & \textbf{27.82} & \textbf{36.14} & \textbf{30.33} & \textbf{34.44} \\
            \hline
            \end{tabular}
    }
    \caption{PSNR($\uparrow$) On UVG at resolution 480x960.}
    \label{tab:uvg480}
    \end{subtable}
    \caption{Video \textbf{reconstruction} results on Bunny and UVG.}
\end{table*}

\begin{table*}[!t]
    \centering
    \tabcolsep=0.22cm
    \begin{tabular}{c|c c c c c c c c c c c}
    \hline
        Video & b-swan & b-trees & boat & b-dance & camel & c-round & c-shadow & cows & dance & dog & avg.  \\ \hline
        NeRV\cite{nerv} & 25.04 & 25.22 & 30.25 & 25.78 & 23.69 & 24.08 & 25.29 & 22.44 & 25.61 & 27.15 & 25.30  \\ 
        DNeRV\cite{dnerv} & 29.84 & 28.73 & 30.52 & 26.58 & 26.24 & 28.50 & 28.88 & 24.44 & 28.42 & 30.64 & 27.79  \\ 
        HNeRV\cite{hnerv} & 29.23 & 28.67 & 32.27 & 31.39 & 25.93 & 28.72 & 31.21 & 24.67 & 28.43 & 30.72 & 28.91  \\ \hline
        Ours & \textbf{32.55} & \textbf{29.76} & \textbf{34.39} & \textbf{32.21} & \textbf{27.26} & \textbf{29.48} & \textbf{35.88} & \textbf{25.08} & \textbf{28.79} & \textbf{33.29} & \textbf{30.36 } \\ 
    \end{tabular}
    \caption{Video \textbf{reconstruction} results on DAVIS, PSNR($\uparrow$) reported.}
    \label{tab:DAVIS_recons}
    \vspace{-3mm}
\end{table*}

\textbf{Datasets.} Extensive experiments are conducted on the Big Buck Bunny~\cite{bunny}, UVG~\cite{uvg} and DAVIS~\cite{DAVIS} datasets. The Bunny dataset has 132 frames with size of $720\times1280$. 
The UVG dataset has 7 videos with resolution of $960\times1920$ and lengths of 600 or 300. 
We select 10 videos from the DAVIS dataset for additional testing, which have a fewer frame number. 
For a fair comparison, we follow the settings in~\cite{hnerv} to crop the Bunny to $640\times1280$ and crop the UVG and DAVIS to $960\times1920$ and also crop a $480\times960$ version of UVG for additional comparison. More details please refer to supplementary material.

\noindent\textbf{Evaluation.} 
We employ PSNR and MS-SSIM~\cite{msssim} as metrics to evaluate video reconstruction quality, and bits per pixel (bpp) as an indicator of video compression performance.
We conduct a comparative analysis between DS-NeRV and other implicit methods, namely NeRV, HNeRV, and DNeRV, in terms of video reconstruction as well as various downstream tasks, including video interpolation and inpainting.
Moreover, we compare video compression performance with existing compression techniques.

\noindent\textbf{Loss Functions.} We only use the L2 loss functions to supervise DS-NeRV, i.e., the decomposition of static codes and dynamic codes is unsupervised.
\begin{equation}
L_2(y_i, \hat{y_i}) = \frac{1}{n} \sum_{i=1}^{n} (y_i - \hat{y}_i)^2
\label{eqa:loss}
\end{equation}
where $y_i$ represents the ground truth and $\hat{y_i}$ is the reconstructed frame.

\noindent\textbf{Implementation Details.} During training, we use the Adan~\cite{adan} optimizer with betas as (0.98,0.92,0.99) and a weight decay of 0.02. 
Moreover, the learning rate is set to $7 \times 10^{-3}$ , and we employ a cosine annealing learning rate schedule with a warm-up ratio of 0.2.
We empirically find that setting the learning rate of static and dynamic code to 10 times the learning rate can achieve better results.
We employ a batch size of 1 on Bunny and DAVIS, while a batch size of 8 on UVG. 
Unless stated otherwise, all models are 3M and trained for 300 epochs. All experiments are performed on the Tesla V100.
More implementation details can be found in the supplementary material.

\subsection{Video Reconstruction}
\label{sec:representation}
We first compare DS-NeRV with other INR methods on Bunny, UVG and DAVIS. 
As shown in Tab.~\ref{tab:bunny}, we evaluate video reconstruction for various model sizes with 300 epochs on Bunny. Remarkably, DS-NeRV achieves impressive video reconstruction quality with a PSNR of 31.20, despite having only 0.35M parameters. 
Furthermore, we evaluate the reconstruction performance with different epochs with a fixed model size of 0.35M, as presented in Tab.~\ref{tab:epochs} and Fig.~\ref{fig:abstract} (right), from which we can see that DS-NeRV converges faster with a higher performance.

We subsequently extend our evaluation on UVG and DAVIS, with qualitative results shown in Fig.~\ref{fig:uvg}. 
DS-NeRV achieves clearer contour reconstruction for the horseshoe in Jockey. Additionally, DS-NeRV captures high-frequency texture details for the leaves in Blackswan, while other methods exhibit noticeable artifacts. This is mainly attributed to our proposed static and dynamic codes representation, which efficiently preserves more detail with compact model size through the utilization of the shared static information and the global temporal-coherence within the video.
More quantitative experimental results are listed in \cref{tab:uvg,tab:uvg480,tab:DAVIS_recons},
which demonstrate a significant improvement achieved by DS-NeRV when compared to other methods, especially in two high-dynamic videos Ready and Jockey. 
We also provide experiment results on the standard UVG of 1080p in supplementary material.

%





 


\begin{figure*}
    \centering
    \includegraphics[width=0.95\textwidth]{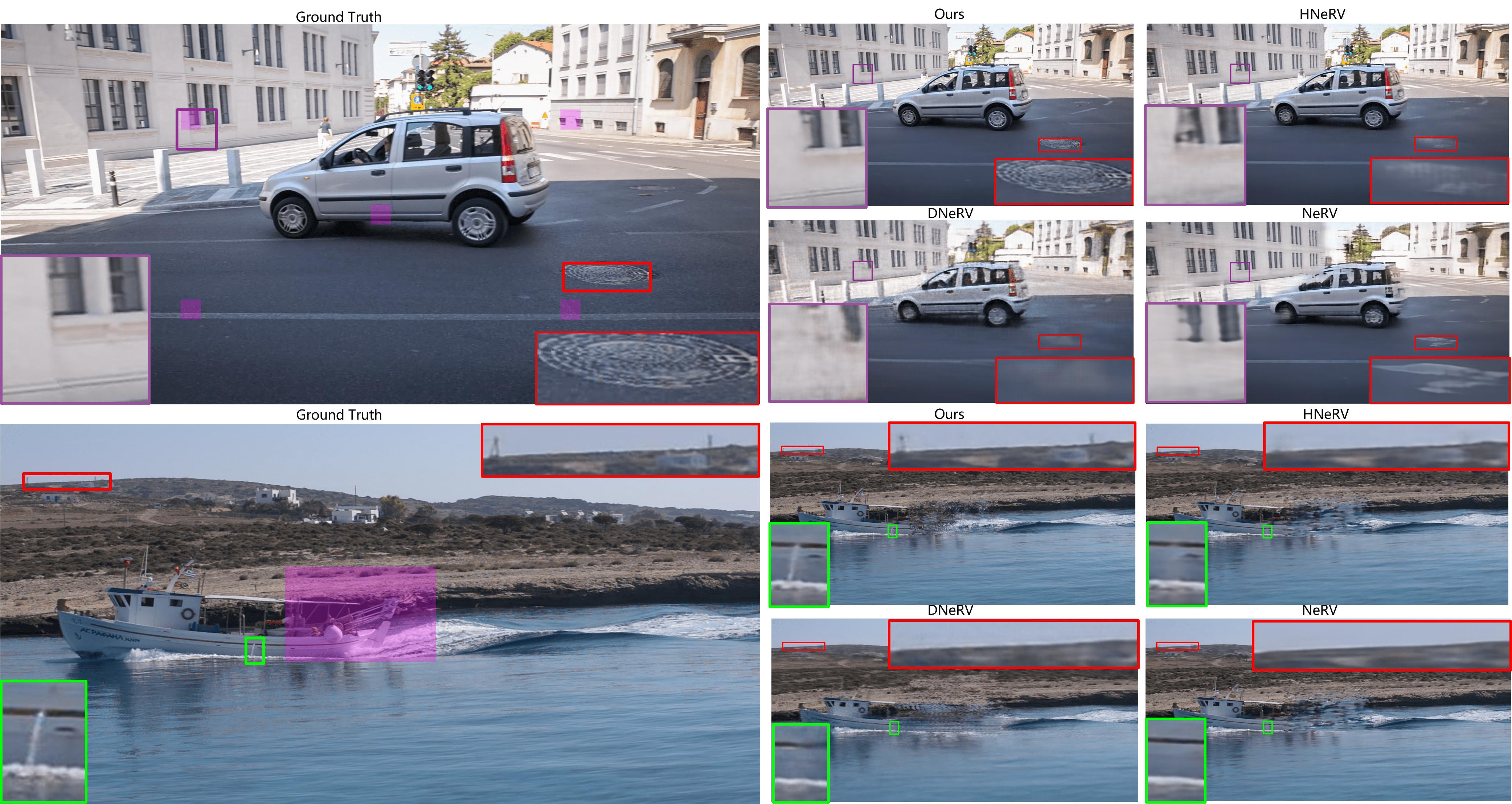}
    \caption{\textbf{Video inpainting results on DAVIS.} (Top) Car-Shadow with 5 masks of width 50.(Bottom) Boat with a central mask with width and height both 1/4 of the video. 
    }
  \label{fig:inpainting}
\end{figure*}

\begin{table*}[!ht]
    \centering
    \begin{subtable}{1.\linewidth}
    \centering
    \tabcolsep=0.22cm
    \begin{tabular}{c|ccccccccccc}
            Video &  b-swan& b-trees &boat &  b-dance & camel & c-round &c-shadow & cows & dance & dog& avg. \\ \hline
            NeRV\cite{nerv} &  24.98& 25.16&30.12&  25.53& 23.65& 24.05&25.17& 22.38& 25.46& 27.05& 25.36\\ 
            DNeRV\cite{dnerv} &  29.52& 28.14&29.52&  25.76& 25.48& 28.00&25.66& 24.05& 27.81& 26.44& 27.03\\ 
             HNeRV\cite{hnerv} &  29.10& 28.67&29.10&  28.67& 26.07& 28.31&30.92& 24.4& 28.44& 30.58& 28.90\\ \hline 
            \textbf{Ours} &  \textbf{32.28}& \textbf{29.58}&\textbf{34.09}&  \textbf{31.50}& \textbf{27.21}& \textbf{29.34}&\textbf{35.35}& \textbf{24.99}& \textbf{28.64}& \textbf{33.03}& \textbf{30.60}\\ 
        \end{tabular}
    
    \caption{PSNR($\uparrow$) On DAVIS with disperse mask.}
    \end{subtable}
    
    \begin{subtable}{1.\linewidth}
    \centering
    \tabcolsep=0.22cm
    \begin{tabular}{c|ccccccccccc}
            Video &  b-swan& b-trees &boat &  b-dance & camel & c-round &c-shadow & cows & dance & dog& avg. \\ \hline
            NeRV\cite{nerv} &  22.72& 22.44 &25.56 &  20.79 & 20.96 & 20.97&22.15 & 20.58& 21.34 & 24.00& 22.15  \\ 
            DNeRV\cite{dnerv} &  26.47& 21.71 &24.74 &  21.96 & 23.10& \textbf{24.41}&\textbf{28.25}& 22.06 & 23.12 & 24.03& 23.99  \\ 
             HNeRV\cite{hnerv} &  26.16& 24.21 &25.96 &  22.20& 22.61 & 22.38&16.32& 21.84 & 22.56 & 26.05& 23.03  \\ \hline 
            \textbf{Ours} &  \textbf{28.33}& \textbf{25.42}&\textbf{27.71}&  \textbf{22.96}& \textbf{23.36}& 24.08&24.89& \textbf{22.71}& \textbf{23.31}& \textbf{27.83}& \textbf{25.06}\\ 
        \end{tabular}
   
    \caption{PSNR($\uparrow$) On DAVIS with central mask.}
    \end{subtable}
    \caption{Video \textbf{inpainting} results on DAVIS.}
    \label{tab:inpaint}
\end{table*}

\subsection{Video Inpainting}
\label{sec:inpainting}
We further investigate the video inpainting on DAVIS. Following the configuration in DNeRV~\cite{dnerv}, we apply masks to the original videos using either five boxes with the size of 50x50 or a central mask with dimensions equal to 1/4 of the width and height of original video. For DS-NeRV and HNeRV, the model is trained on the masked video, while DNeRV is trained on the original video according to their setting. All methods are tested on the masked videos and the qualitative details of inpainted frames are shown in Fig.~\ref{fig:inpainting}. 
Note that the windows in Car-shadow (Top) and water flow in Boat (Bottom) are masked in some certain video frames.
DS-NeRV almost perfectly inpaints the masked areas thanks to the utilization of global temporal coherence and its capacity to learn and then fill the masked areas using information visible in other frames.
Moreover, DS-NeRV successfully reconstruct high-frequency details, such as the manhole cover in Car-shadow and the distant electric wire tower in Boat.
More quantitative experimental results are presented in the Tab.~\ref{tab:inpaint}, demonstrating the superiority of our proposed DS-NeRV over other methods.

\subsection{Video Interpolation}
\label{sec:interpolation}
\begin{figure*}
    \centering
    \includegraphics[width=0.95\textwidth]{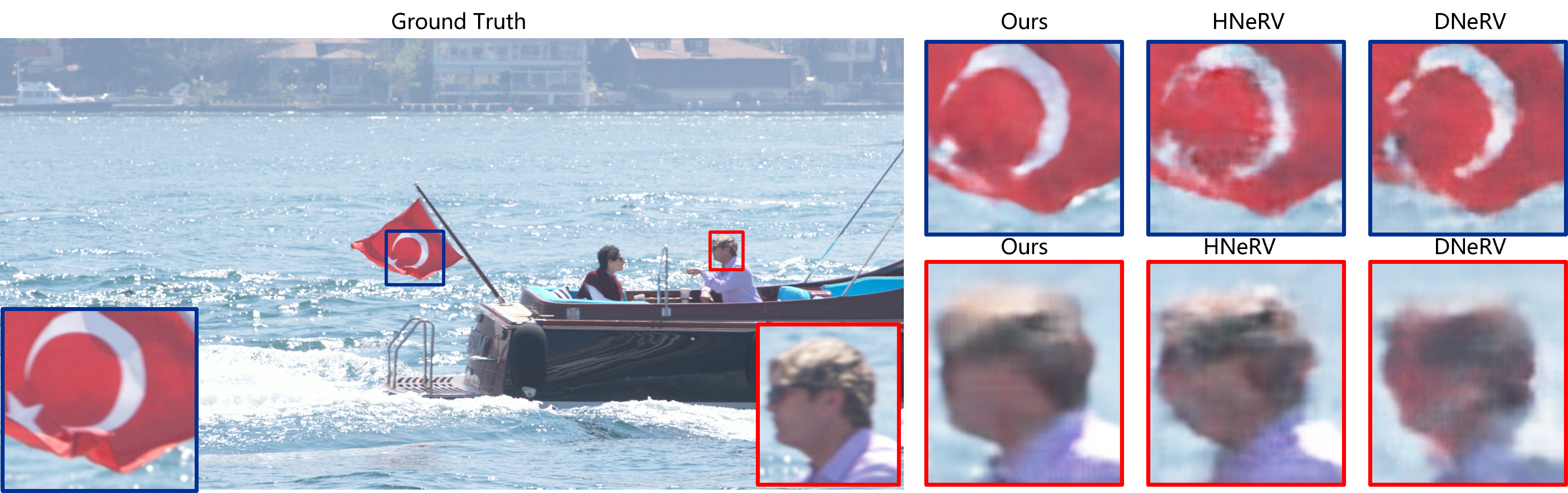}
    \caption{\textbf{Video interpolation results on UVG, interpolated frame shown above.} (a) Ours. (b) HNeRV. (c) DNeRV. 
    }
  \label{fig:interpolation}
\end{figure*}

We use even-numbered frames from the video as the training set and odd-numbered frames as the test set to conduct the interpolation experiment. During testing, DS-NeRV utilizes trained interpolated static and dynamic codes as inputs, in this way, our method can naturally generalize to frames that are not seen in the training set. 
The way we conduct the test is similar to typically video interpolation task~\cite{transformerinterpolation,interframe-attention-interp}, where the frames to be interpolated are not visible during training and testing.
However, HNeRV and DNeRV use test frames itself as input during testing to obtain embeddings, which are subsequently used to generate corresponding ground truth, which is not practical because the test frames are typically unknown.
The quantitative results on the training and test sets are shown in Tab.~\ref{tab:interpolation}, which demonstrates the superior performance of DS-NeRV on the training set compared to existing methods. Furthermore, DS-NeRV also achieves comparable performance on the test set even without seeing the ground truth during testing.
The qualitative results on interpolation can be found in Fig.~\ref{fig:interpolation}, where DS-NeRV achieves better reconstruction of the flag pattern and exhibits clearer contour in the human head region.

\begin{table*}[t]
    \centering
    \tabcolsep=0.112cm
        \resizebox{1\linewidth}{!}{
        \begin{tabular}{c|ccccccc|c}
        \hline
            video & Beauty & Bosph & Honey & Jockey & Ready & Shake & Yacht & avg. \\ \hline
            HNeRV\cite{hnerv} & 34.02/31.26& 34.69/34.54& 39.26/39.10& 32.58/22.86& 26.25/20.51& \textbf{34.91}/32.79& 29.20/27.41& 32.99/29.78\\ 
            DNeRV\cite{dnerv} & 33.46/\textbf{32.48}& 30.96/30.77& 38.55/38.36&32.22/\textbf{29.79}&25.78/\textbf{24.29}& 34.41/\textbf{33.34}& 26.37/25.96& 31.68\textbf{30.71}\\ \hline
            Ours & \textbf{34.08}/31.84& \textbf{34.96}/\textbf{34.82}& \textbf{39.48}/\textbf{39.27}& \textbf{33.60}/22.96& \textbf{27.48}/21.26& 34.54/33.17& \textbf{29.55}/\textbf{27.52}& \textbf{33.30}/30.09\\ \hline
        \end{tabular}
        }
    \caption{Video \textbf{interpolation} results on UVG with train/test split, PSNR($\uparrow$) reported.}
    \label{tab:interpolation}
    \vspace{-2mm}
\end{table*}
\begin{figure}
    \centering
    \includegraphics[width=0.48\textwidth]{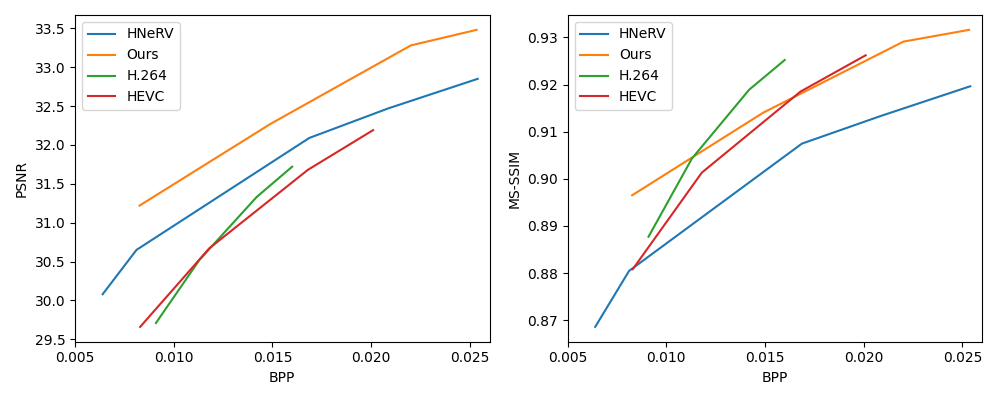}
    \caption{Compression performance on UVG dataset.}
  \label{fig:compression}
  \vspace{-4mm}
\end{figure}
\subsection{Video Compression}
\label{sec:compression}
We follow the process in HNeRV to compress the model through model quantization, model pruning, and entropy coding. We compare DS-NeRV with H.264~\cite{h264}, HEVC~\cite{hevc}, NeRV~\cite{nerv} and HNeRV~\cite{hnerv}. We present the results of video compression in Fig.~\ref{fig:compression}. From the figure we can see that DS-NeRV surpasses HNeRV, exhibiting significant improvements. Additionally, in many cases, our method outperforms traditional methods such as H.264 and HEVC, achieving superior performance. The experimental results validate the effectiveness of our compression strategy.
\subsection{Ablation Studies}
\label{sec:abla}
\noindent\textbf{Static/Dynamic codes.}
To evaluate the effectiveness of the static and dynamic code designs and the impact of their lengths on video reconstruction, we conduct ablation experiments on Jockey and Honey from the UVG. Jockey exhibits strong dynamics, while Honey features nearly static video frames. 

The results of the ablation experiments are presented in the \cref{tab:abla-codelength}. The results demonstrate the varying effects of different combinations of static and dynamic code lengths on videos with different levels of dynamics. For Jockey, increasing the length of the dynamic codes gradually improves the video quality, while the effect is less pronounced for Honey. When one of the lengths is set to 0, indicating the absence of the corresponding code, it further confirms that both static and dynamic codes are essential elements for achieving high-quality reconstruction, highlighting the necessity of their collaboration. Appropriately setting the lengths under a certain model size enables the model to fully utilize and compress the redundant static information contained in static codes and the dynamic information in dynamic codes.
The static and dynamic parts of the video, decoded from the corresponding static and dynamic codes, are shown in the Fig.~\ref{fig:abstract} (Left). More ablation results can be found in the supplementary material.


\begin{table}[t]
\centering
    \tabcolsep=0.05cm
    \begin{tabular}{c|cccc}
        \hline
        $t_s$$\backslash$$t_d$ & 0 & 100 & 200 & 300 \\ \hline
        0 & \textcolor{red}{29.91}/\textcolor{blue}{38.80} & \textcolor{red}{30.78}/\textcolor{blue}{39.49} & \textcolor{red}{31.10}/\textcolor{blue}{39.41} & \textcolor{red}{31.62}/\textcolor{blue}{39.21} \\ \hline
        30 & \textcolor{red}{28.68}/\textcolor{blue}{39.13} & \textcolor{red}{30.87}/\textcolor{blue}{39.53} & \textcolor{red}{32.16}/\textcolor{blue}{39.52} & \textcolor{red}{32.56}/\textcolor{blue}{39.43} \\ \hline
        60 & \textcolor{red}{29.91}/\textcolor{blue}{39.23} & \textcolor{red}{31.04}/\textcolor{blue}{39.53} & \textcolor{red}{32.25}/\textcolor{blue}{39.52} & \textcolor{red}{32.75}/\textcolor{blue}{39.43} \\ \hline
        90 & \textcolor{red}{30.88}/\textcolor{blue}{39.38} & \textcolor{red}{31.19}/\textcolor{blue}{39.54} & \textcolor{red}{32.23}/\textcolor{blue}{39.52} & \textcolor{red}{32.69}/\textcolor{blue}{39.42} \\ \hline
    \end{tabular}
    \label{tab:abla-codelength}
    \caption{Ablation study for codes length on \textcolor{red}{Jockey}/\textcolor{blue}{Honey}. The (0,0) combination refers to vanilla NeRV.}
\end{table}


\section{Conclusion}
In this paper, we propose DS-NeRV, a novel INR for video, that decomposes the video into sparse, learnable static and dynamic codes. 
By computing a weighted sum of the static codes and interpolating the dynamic codes, DS-NeRV effectively utilizes the redundancy of static information in videos and models global temporal-coherent dynamic information.
According to our extensive experiment, DS-NeRV outperforms the state-of-the-art methods in many downstream tasks.

\textbf{Acknowledgements.} This work is support in part by  the National Science Fund for Distinguished Young Scholars (No. 62325208), in part by NSFC Joint Funds (No. U2001204), and in part by the National Natural Science Foundation of China (No. 62232002, 62072330).
{
    \small
    \bibliographystyle{ieeenat_fullname}
    \bibliography{main}

\begin{thebibliography}{55}
\providecommand{\natexlab}[1]{#1}
\providecommand{\url}[1]{\texttt{#1}}
\expandafter\ifx\csname urlstyle\endcsname\relax
  \providecommand{\doi}[1]{doi: #1}\else
  \providecommand{\doi}{doi: \begingroup \urlstyle{rm}\Url}\fi

\bibitem[cis()]{cisco}
Cisco: Service provider solutions.
\newblock \url{https://www.cisco.com/c/en/us/solutions/service-provider/index.html\#~products-and-solutions}.

\bibitem[san()]{sandvine}
2023 global internet phenomena report.
\newblock \url{https://www.sandvine.com/phenomena}.

\bibitem[Agustsson et~al.(2020)Agustsson, Minnen, Johnston, Balle, Hwang, and Toderici]{scale-space}
Eirikur Agustsson, David Minnen, Nick Johnston, Johannes Balle, Sung~Jin Hwang, and George Toderici.
\newblock Scale-space flow for end-to-end optimized video compression.
\newblock In \emph{Proceedings of the IEEE/CVF Conference on Computer Vision and Pattern Recognition}, pages 8503--8512, 2020.

\bibitem[Bai et~al.(2023)Bai, Dong, Wang, and Yuan]{psnerv}
Yunpeng Bai, Chao Dong, Cairong Wang, and Chun Yuan.
\newblock Ps-nerv: Patch-wise stylized neural representations for videos.
\newblock In \emph{2023 IEEE International Conference on Image Processing (ICIP)}, pages 41--45. IEEE, 2023.

\bibitem[Barron et~al.(2021)Barron, Mildenhall, Tancik, Hedman, Martin-Brualla, and Srinivasan]{mipnerf}
Jonathan~T Barron, Ben Mildenhall, Matthew Tancik, Peter Hedman, Ricardo Martin-Brualla, and Pratul~P Srinivasan.
\newblock Mip-nerf: A multiscale representation for anti-aliasing neural radiance fields.
\newblock In \emph{Proceedings of the IEEE/CVF International Conference on Computer Vision}, pages 5855--5864, 2021.

\bibitem[Bian et~al.(2023)Bian, Wang, Li, Bian, and Prisacariu]{nope}
Wenjing Bian, Zirui Wang, Kejie Li, Jia-Wang Bian, and Victor~Adrian Prisacariu.
\newblock Nope-nerf: Optimising neural radiance field with no pose prior.
\newblock In \emph{Proceedings of the IEEE/CVF Conference on Computer Vision and Pattern Recognition}, pages 4160--4169, 2023.

\bibitem[Bojanowski et~al.(2017)Bojanowski, Joulin, Lopez-Paz, and Szlam]{glo}
Piotr Bojanowski, Armand Joulin, David Lopez-Paz, and Arthur Szlam.
\newblock Optimizing the latent space of generative networks.
\newblock \emph{arXiv preprint arXiv:1707.05776}, 2017.

\bibitem[Chen et~al.(2022{\natexlab{a}})Chen, Xu, Geiger, Yu, and Su]{tensorf}
Anpei Chen, Zexiang Xu, Andreas Geiger, Jingyi Yu, and Hao Su.
\newblock Tensorf: Tensorial radiance fields.
\newblock In \emph{European Conference on Computer Vision}, pages 333--350. Springer, 2022{\natexlab{a}}.

\bibitem[Chen et~al.(2021)Chen, He, Wang, Ren, Lim, and Shrivastava]{nerv}
Hao Chen, Bo He, Hanyu Wang, Yixuan Ren, Ser~Nam Lim, and Abhinav Shrivastava.
\newblock Nerv: Neural representations for videos.
\newblock \emph{Advances in Neural Information Processing Systems}, 34:\penalty0 21557--21568, 2021.

\bibitem[Chen et~al.(2022{\natexlab{b}})Chen, Gwilliam, He, Lim, and Shrivastava]{cnerv}
Hao Chen, Matt Gwilliam, Bo He, Ser-Nam Lim, and Abhinav Shrivastava.
\newblock Cnerv: Content-adaptive neural representation for visual data.
\newblock \emph{arXiv preprint arXiv:2211.10421}, 2022{\natexlab{b}}.

\bibitem[Chen et~al.(2023)Chen, Gwilliam, Lim, and Shrivastava]{hnerv}
Hao Chen, Matthew Gwilliam, Ser-Nam Lim, and Abhinav Shrivastava.
\newblock Hnerv: A hybrid neural representation for videos.
\newblock In \emph{Proceedings of the IEEE/CVF Conference on Computer Vision and Pattern Recognition}, pages 10270--10279, 2023.

\bibitem[Chen et~al.(2017)Chen, Liu, Shen, Yue, Cao, and Ma]{deepcoder}
Tong Chen, Haojie Liu, Qiu Shen, Tao Yue, Xun Cao, and Zhan Ma.
\newblock Deepcoder: A deep neural network based video compression.
\newblock In \emph{IEEE Visual Communications and Image Processing Conference}, 2017.

\bibitem[Chen et~al.(2018)Chen, Murherjee, Han, Grange, Xu, Liu, Parker, Chen, Su, Joshi, et~al.]{av1}
Yue Chen, Debargha Murherjee, Jingning Han, Adrian Grange, Yaowu Xu, Zoe Liu, Sarah Parker, Cheng Chen, Hui Su, Urvang Joshi, et~al.
\newblock An overview of core coding tools in the av1 video codec.
\newblock In \emph{2018 picture coding symposium (PCS)}, pages 41--45. IEEE, 2018.

\bibitem[Chen et~al.(2019)Chen, He, Jin, and Wu]{pcnn}
Zhibo Chen, Tianyu He, Xin Jin, and Feng Wu.
\newblock Learning for video compression.
\newblock \emph{IEEE Transactions on Circuits and Systems for Video Technology}, 30\penalty0 (2):\penalty0 566--576, 2019.

\bibitem[Dosovitskiy et~al.(2020)Dosovitskiy, Beyer, Kolesnikov, Weissenborn, Zhai, Unterthiner, Dehghani, Minderer, Heigold, Gelly, et~al.]{vit}
Alexey Dosovitskiy, Lucas Beyer, Alexander Kolesnikov, Dirk Weissenborn, Xiaohua Zhai, Thomas Unterthiner, Mostafa Dehghani, Matthias Minderer, Georg Heigold, Sylvain Gelly, et~al.
\newblock An image is worth 16x16 words: Transformers for image recognition at scale.
\newblock \emph{arXiv preprint arXiv:2010.11929}, 2020.

\bibitem[Du et~al.(2021)Du, Zhang, Yu, Tenenbaum, and Wu]{nerfflow}
Yilun Du, Yinan Zhang, Hong-Xing Yu, Joshua~B Tenenbaum, and Jiajun Wu.
\newblock Neural radiance flow for 4d view synthesis and video processing.
\newblock In \emph{2021 IEEE/CVF International Conference on Computer Vision (ICCV)}, pages 14304--14314. IEEE Computer Society, 2021.

\bibitem[Fridovich-Keil et~al.(2023)Fridovich-Keil, Meanti, Warburg, Recht, and Kanazawa]{kplanes}
Sara Fridovich-Keil, Giacomo Meanti, Frederik~Rahb{\ae}k Warburg, Benjamin Recht, and Angjoo Kanazawa.
\newblock K-planes: Explicit radiance fields in space, time, and appearance.
\newblock In \emph{Proceedings of the IEEE/CVF Conference on Computer Vision and Pattern Recognition (CVPR)}, pages 12479--12488, 2023.

\bibitem[He et~al.(2023)He, Yang, Wang, Wu, Chen, Huang, Ren, Lim, and Shrivastava]{d-nerv}
Bo He, Xitong Yang, Hanyu Wang, Zuxuan Wu, Hao Chen, Shuaiyi Huang, Yixuan Ren, Ser-Nam Lim, and Abhinav Shrivastava.
\newblock Towards scalable neural representation for diverse videos.
\newblock In \emph{Proceedings of the IEEE/CVF Conference on Computer Vision and Pattern Recognition}, pages 6132--6142, 2023.

\bibitem[Jiang et~al.(2022)Jiang, Jiang, Grauman, and Zhu]{few-view}
Hanwen Jiang, Zhenyu Jiang, Kristen Grauman, and Yuke Zhu.
\newblock Few-view object reconstruction with unknown categories and camera poses.
\newblock \emph{arXiv preprint arXiv:2212.04492}, 2022.

\bibitem[Kim et~al.(2022)Kim, Yu, Lee, and Shin]{nvp}
Subin Kim, Sihyun Yu, Jaeho Lee, and Jinwoo Shin.
\newblock Scalable neural video representations with learnable positional features.
\newblock \emph{Advances in Neural Information Processing Systems}, 35:\penalty0 12718--12731, 2022.

\bibitem[Lee et~al.(2022)Lee, Rho, Ko, and Park]{ffnerv}
Joo~Chan Lee, Daniel Rho, Jong~Hwan Ko, and Eunbyung Park.
\newblock Ffnerv: Flow-guided frame-wise neural representations for videos.
\newblock \emph{arXiv preprint arXiv:2212.12294}, 2022.

\bibitem[LEGALL(1993)]{mpeg}
D LEGALL.
\newblock A video compression standard for multimedia applications.
\newblock \emph{Commun. ACM}, 34:\penalty0 226--252, 1993.

\bibitem[Li et~al.(2021)Li, Li, and Lu]{dcvc}
Jiahao Li, Bin Li, and Yan Lu.
\newblock Deep contextual video compression.
\newblock \emph{Advances in Neural Information Processing Systems}, 34:\penalty0 18114--18125, 2021.

\bibitem[Li et~al.(2022{\natexlab{a}})Li, Slavcheva, Zollhoefer, Green, Lassner, Kim, Schmidt, Lovegrove, Goesele, Newcombe, et~al.]{n3dv}
Tianye Li, Mira Slavcheva, Michael Zollhoefer, Simon Green, Christoph Lassner, Changil Kim, Tanner Schmidt, Steven Lovegrove, Michael Goesele, Richard Newcombe, et~al.
\newblock Neural 3d video synthesis from multi-view video.
\newblock In \emph{Proceedings of the IEEE/CVF Conference on Computer Vision and Pattern Recognition}, pages 5521--5531, 2022{\natexlab{a}}.

\bibitem[Li et~al.(2022{\natexlab{b}})Li, Wang, Pi, Xu, Mei, and Liu]{enerv}
Zizhang Li, Mengmeng Wang, Huaijin Pi, Kechun Xu, Jianbiao Mei, and Yong Liu.
\newblock E-nerv: Expedite neural video representation with disentangled spatial-temporal context.
\newblock In \emph{European Conference on Computer Vision}, pages 267--284. Springer, 2022{\natexlab{b}}.

\bibitem[Lin et~al.(2020)Lin, Liu, Li, and Wu]{m-lvc}
Jianping Lin, Dong Liu, Houqiang Li, and Feng Wu.
\newblock M-lvc: Multiple frames prediction for learned video compression.
\newblock In \emph{Proceedings of the IEEE/CVF Conference on Computer Vision and Pattern Recognition}, pages 3546--3554, 2020.

\bibitem[Liu et~al.(2016)Liu, Yu, Gao, Chen, Ji, and Wang]{cu}
Zhenyu Liu, Xianyu Yu, Yuan Gao, Shaolin Chen, Xiangyang Ji, and Dongsheng Wang.
\newblock Cu partition mode decision for hevc hardwired intra encoder using convolution neural network.
\newblock \emph{IEEE Transactions on Image Processing}, 25\penalty0 (11):\penalty0 5088--5103, 2016.

\bibitem[Lu et~al.(2019)Lu, Ouyang, Xu, Zhang, Cai, and Gao]{dvc}
Guo Lu, Wanli Ouyang, Dong Xu, Xiaoyun Zhang, Chunlei Cai, and Zhiyong Gao.
\newblock Dvc: An end-to-end deep video compression framework.
\newblock In \emph{Proceedings of the IEEE/CVF Conference on Computer Vision and Pattern Recognition}, pages 11006--11015, 2019.

\bibitem[Lu et~al.(2022)Lu, Wu, Lin, Lu, and Jia]{transformerinterpolation}
Liying Lu, Ruizheng Wu, Huaijia Lin, Jiangbo Lu, and Jiaya Jia.
\newblock Video frame interpolation with transformer.
\newblock In \emph{Proceedings of the IEEE/CVF Conference on Computer Vision and Pattern Recognition}, pages 3532--3542, 2022.

\bibitem[Maiya et~al.(2023)Maiya, Girish, Ehrlich, Wang, Lee, Poirson, Wu, Wang, and Shrivastava]{nirvana}
Shishira~R Maiya, Sharath Girish, Max Ehrlich, Hanyu Wang, Kwot~Sin Lee, Patrick Poirson, Pengxiang Wu, Chen Wang, and Abhinav Shrivastava.
\newblock Nirvana: Neural implicit representations of videos with adaptive networks and autoregressive patch-wise modeling.
\newblock In \emph{Proceedings of the IEEE/CVF Conference on Computer Vision and Pattern Recognition}, pages 14378--14387, 2023.

\bibitem[Martin-Brualla et~al.(2021)Martin-Brualla, Radwan, Sajjadi, Barron, Dosovitskiy, and Duckworth]{nerf-w}
Ricardo Martin-Brualla, Noha Radwan, Mehdi~SM Sajjadi, Jonathan~T Barron, Alexey Dosovitskiy, and Daniel Duckworth.
\newblock Nerf in the wild: Neural radiance fields for unconstrained photo collections.
\newblock In \emph{Proceedings of the IEEE/CVF Conference on Computer Vision and Pattern Recognition}, pages 7210--7219, 2021.

\bibitem[Mercat et~al.(2020)Mercat, Viitanen, and Vanne]{uvg}
Alexandre Mercat, Marko Viitanen, and Jarno Vanne.
\newblock Uvg dataset: 50/120fps 4k sequences for video codec analysis and development.
\newblock In \emph{Proceedings of the 11th ACM Multimedia Systems Conference}, pages 297--302, 2020.

\bibitem[Mescheder et~al.(2019)Mescheder, Oechsle, Niemeyer, Nowozin, and Geiger]{occupancy}
Lars Mescheder, Michael Oechsle, Michael Niemeyer, Sebastian Nowozin, and Andreas Geiger.
\newblock Occupancy networks: Learning 3d reconstruction in function space.
\newblock In \emph{Proceedings of the IEEE/CVF conference on computer vision and pattern recognition}, pages 4460--4470, 2019.

\bibitem[Mildenhall et~al.(2021)Mildenhall, Srinivasan, Tancik, Barron, Ramamoorthi, and Ng]{nerf}
Ben Mildenhall, Pratul~P Srinivasan, Matthew Tancik, Jonathan~T Barron, Ravi Ramamoorthi, and Ren Ng.
\newblock Nerf: Representing scenes as neural radiance fields for view synthesis.
\newblock \emph{Communications of the ACM}, 65\penalty0 (1):\penalty0 99--106, 2021.

\bibitem[M{\"u}ller et~al.(2022)M{\"u}ller, Evans, Schied, and Keller]{ingp}
Thomas M{\"u}ller, Alex Evans, Christoph Schied, and Alexander Keller.
\newblock Instant neural graphics primitives with a multiresolution hash encoding.
\newblock \emph{ACM Transactions on Graphics (ToG)}, 41\penalty0 (4):\penalty0 1--15, 2022.

\bibitem[Park et~al.(2019)Park, Florence, Straub, Newcombe, and Lovegrove]{deepsdf}
Jeong~Joon Park, Peter Florence, Julian Straub, Richard Newcombe, and Steven Lovegrove.
\newblock Deepsdf: Learning continuous signed distance functions for shape representation.
\newblock In \emph{Proceedings of the IEEE/CVF conference on computer vision and pattern recognition}, pages 165--174, 2019.

\bibitem[Perazzi et~al.(2016)Perazzi, Pont-Tuset, McWilliams, Van~Gool, Gross, and Sorkine-Hornung]{DAVIS}
Federico Perazzi, Jordi Pont-Tuset, Brian McWilliams, Luc Van~Gool, Markus Gross, and Alexander Sorkine-Hornung.
\newblock A benchmark dataset and evaluation methodology for video object segmentation.
\newblock In \emph{Proceedings of the IEEE conference on computer vision and pattern recognition}, pages 724--732, 2016.

\bibitem[Pumarola et~al.(2021)Pumarola, Corona, Pons-Moll, and Moreno-Noguer]{dnerf}
Albert Pumarola, Enric Corona, Gerard Pons-Moll, and Francesc Moreno-Noguer.
\newblock D-nerf: Neural radiance fields for dynamic scenes.
\newblock In \emph{Proceedings of the IEEE/CVF Conference on Computer Vision and Pattern Recognition}, pages 10318--10327, 2021.

\bibitem[Rippel et~al.(2021)Rippel, Anderson, Tatwawadi, Nair, Lytle, and Bourdev]{elf}
Oren Rippel, Alexander~G Anderson, Kedar Tatwawadi, Sanjay Nair, Craig Lytle, and Lubomir Bourdev.
\newblock Elf-vc: Efficient learned flexible-rate video coding.
\newblock In \emph{Proceedings of the IEEE/CVF International Conference on Computer Vision}, pages 14479--14488, 2021.

\bibitem[Roosendaal(2008)]{bunny}
Ton Roosendaal.
\newblock Big buck bunny.
\newblock In \emph{ACM SIGGRAPH ASIA 2008 computer animation festival}, pages 62--62. 2008.

\bibitem[Sitzmann et~al.(2020)Sitzmann, Martel, Bergman, Lindell, and Wetzstein]{siren}
Vincent Sitzmann, Julien Martel, Alexander Bergman, David Lindell, and Gordon Wetzstein.
\newblock Implicit neural representations with periodic activation functions.
\newblock \emph{Advances in neural information processing systems}, 33:\penalty0 7462--7473, 2020.

\bibitem[Song et~al.(2017)Song, Liu, Li, and Wu]{entropycoding}
Rui Song, Dong Liu, Houqiang Li, and Feng Wu.
\newblock Neural network-based arithmetic coding of intra prediction modes in hevc, 2017.

\bibitem[Sullivan et~al.(2012)Sullivan, Ohm, Han, and Wiegand]{hevc}
Gary~J Sullivan, Jens-Rainer Ohm, Woo-Jin Han, and Thomas Wiegand.
\newblock Overview of the high efficiency video coding (hevc) standard.
\newblock \emph{IEEE Transactions on circuits and systems for video technology}, 22\penalty0 (12):\penalty0 1649--1668, 2012.

\bibitem[Tancik et~al.(2022)Tancik, Casser, Yan, Pradhan, Mildenhall, Srinivasan, Barron, and Kretzschmar]{blocknerf}
Matthew Tancik, Vincent Casser, Xinchen Yan, Sabeek Pradhan, Ben Mildenhall, Pratul~P Srinivasan, Jonathan~T Barron, and Henrik Kretzschmar.
\newblock Block-nerf: Scalable large scene neural view synthesis.
\newblock In \emph{Proceedings of the IEEE/CVF Conference on Computer Vision and Pattern Recognition}, pages 8248--8258, 2022.

\bibitem[Tretschk et~al.(2021)Tretschk, Tewari, Golyanik, Zollh{\"o}fer, Lassner, and Theobalt]{nrnerf}
Edgar Tretschk, Ayush Tewari, Vladislav Golyanik, Michael Zollh{\"o}fer, Christoph Lassner, and Christian Theobalt.
\newblock Non-rigid neural radiance fields: Reconstruction and novel view synthesis of a dynamic scene from monocular video.
\newblock In \emph{Proceedings of the IEEE/CVF International Conference on Computer Vision}, pages 12959--12970, 2021.

\bibitem[Wang et~al.(2003)Wang, Simoncelli, and Bovik]{msssim}
Zhou Wang, Eero~P Simoncelli, and Alan~C Bovik.
\newblock Multiscale structural similarity for image quality assessment.
\newblock In \emph{The Thrity-Seventh Asilomar Conference on Signals, Systems \& Computers, 2003}, pages 1398--1402. Ieee, 2003.

\bibitem[Wiegand et~al.(2003)Wiegand, Sullivan, Bjontegaard, and Luthra]{h264}
Thomas Wiegand, Gary~J Sullivan, Gisle Bjontegaard, and Ajay Luthra.
\newblock Overview of the h. 264/avc video coding standard.
\newblock \emph{IEEE Transactions on circuits and systems for video technology}, 13\penalty0 (7):\penalty0 560--576, 2003.

\bibitem[Woo et~al.(2018)Woo, Park, Lee, and Kweon]{cbam}
Sanghyun Woo, Jongchan Park, Joon-Young Lee, and In~So Kweon.
\newblock Cbam: Convolutional block attention module.
\newblock In \emph{Proceedings of the European conference on computer vision (ECCV)}, pages 3--19, 2018.

\bibitem[Wu et~al.(2018)Wu, Singhal, and Krahenbuhl]{wu_eccv}
Chao-Yuan Wu, Nayan Singhal, and Philipp Krahenbuhl.
\newblock Video compression through image interpolation.
\newblock In \emph{Proceedings of the European Conference on Computer Vision (ECCV)}, 2018.

\bibitem[Wu et~al.(2019)Wu, Donahue, Balduzzi, Simonyan, and Lillicrap]{latent_optimisation}
Yan Wu, Jeff Donahue, David Balduzzi, Karen Simonyan, and Timothy Lillicrap.
\newblock Logan: Latent optimisation for generative adversarial networks.
\newblock \emph{arXiv preprint arXiv:1912.00953}, 2019.

\bibitem[Xie et~al.(2022)Xie, Zhou, Li, Lin, and Yan]{adan}
Xingyu Xie, Pan Zhou, Huan Li, Zhouchen Lin, and Shuicheng Yan.
\newblock Adan: Adaptive nesterov momentum algorithm for faster optimizing deep models.
\newblock \emph{arXiv preprint arXiv:2208.06677}, 2022.

\bibitem[Yang et~al.(2020)Yang, Mentzer, Gool, and Timofte]{hlvc}
Ren Yang, Fabian Mentzer, Luc~Van Gool, and Radu Timofte.
\newblock Learning for video compression with hierarchical quality and recurrent enhancement.
\newblock In \emph{Proceedings of the IEEE/CVF Conference on Computer Vision and Pattern Recognition}, pages 6628--6637, 2020.

\bibitem[Zhang et~al.(2023)Zhang, Zhu, Wang, Chen, Wu, and Wang]{interframe-attention-interp}
Guozhen Zhang, Yuhan Zhu, Haonan Wang, Youxin Chen, Gangshan Wu, and Limin Wang.
\newblock Extracting motion and appearance via inter-frame attention for efficient video frame interpolation.
\newblock In \emph{Proceedings of the IEEE/CVF Conference on Computer Vision and Pattern Recognition}, pages 5682--5692, 2023.

\bibitem[Zhao et~al.(2023)Zhao, Asif, and Ma]{dnerv}
Qi Zhao, M~Salman Asif, and Zhan Ma.
\newblock Dnerv: Modeling inherent dynamics via difference neural representation for videos.
\newblock In \emph{Proceedings of the IEEE/CVF Conference on Computer Vision and Pattern Recognition}, pages 2031--2040, 2023.

\bibitem[Zhou and Kr{\"a}henb{\"u}hl(2022)]{cross}
Brady Zhou and Philipp Kr{\"a}henb{\"u}hl.
\newblock Cross-view transformers for real-time map-view semantic segmentation.
\newblock In \emph{Proceedings of the IEEE/CVF conference on computer vision and pattern recognition}, pages 13760--13769, 2022.

\end{thebibliography}
}
\clearpage

\maketitlesupplementary
\appendix
\section{Video Modeling Details}
\label{sec:supp_modeling}
\textbf{Static Codes.} In our implementation, for simplicity, we determine the static codes length $l_s$ by directly specifying it, bypassing the formula that contains the sampling rate $r_s$.
Typically the value of $l_s$ is the sum of a factor of the video length T, denoted as $T_f$, and 1 (\ie~$l_s = T_f + 1$), where the additional 1 refers to the last static code placed on the final frame of the video. This ensures that a fixed interval $z_s$ between most static codes (\ie~$z_s=T/l_s$), excluding the last two.

\begin{figure}
    \centering
    \includegraphics[width=1.0\linewidth]{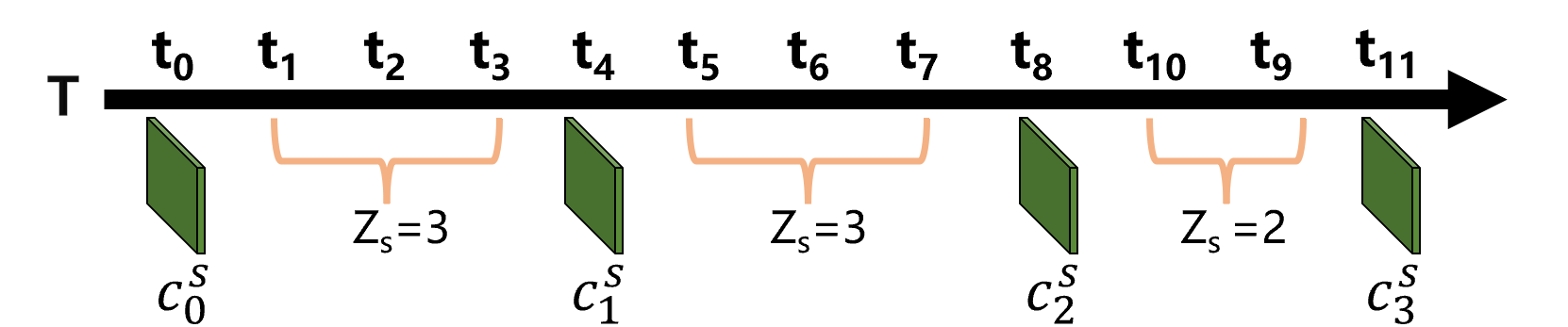}
    \caption{Static Codes Sampling.}
    \label{fig:static_sample}
\end{figure}

Specifically, as shown in Fig.~\ref{fig:static_sample}, given $T=12$, we choose the $T_f$ as $3$, thus the length of the static codes $l_s$ is computed as $l_s = T_f +1 = 4$. The intervals of the first three static codes is set to $T/l_s=3$, while the interval of the last two static codes is 2.

\noindent \textbf{Dynamic Codes.} Similar to the static codes, we also directly specify the value of $l_d$ to determine the length of the dynamic codes, thus bypassing the calculation containing the sampling rate $r_d$. Since we do not employ the sampling method of weighted sum used in static code sampling but use interpolation when sampling dynamic code corresponding to frame index $t$, so the dynamic codes length $l_d$  has greater flexibility and can be set freely. Typically, we set the $l_d$ to be approximately half of the video length $T$. This approach not only avoids the storage overhead of saving a dynamic code for each frame but also ensures sufficient dynamic information for subsequent interpolation operations.

\section{Additional Setup}
\subsection{Datasets}
We conduct experiments using 7 videos from UVG: Beauty, Bosphorus, HoneyBee, Jockey, ReadySetGo, ShakeNDry, and YachtRide. Except for ShakeNDry, which consists of 300 frames, the remaining videos contain 600 frames each. The resolution of all the videos is  $960\times1920$. We also select 10 videos in DAVIS to conduct experiments. The chosen videos include Blackswan, Bmx-trees, Boat, Breakdance, Camel, Car-roundabout, Car-shadow, Cows, Dance, and Dog. These videos have a relatively small number of frames, presenting significant challenges for our experiments.
\subsection{Implementation.}
For NeRV and HNeRV, we conduct experiments using their open-source implementations. As for DNeRV, we develop our implementation based on the open-source E-NeRV code. When comparing the model sizes between DS-NeRV and other implicit methods, the total size of HNeRV comprises the sum of its embedding and decoder, whereas in the case of DNeRV, the total size is calculated by summing the diff embedding, content embedding, and the decoder. As for DS-NeRV, the total size includes the sum of the static and dynamic codes as well as the fusion decoder. 

In our typical implementation, for a $960\times1920\times3$ video frame, we configure the dimensions of each static code as $4\times8\times64$. The dimensions of each dynamic code is set to $20\times40\times2$. The lengths of the static and dynamic codes, while depending on the extent of changes in video dynamics, are significantly shorter than the original video length. In a few videos with strong dynamic changes, such as Ready and Jockey, we fine-tune the dynamic codes dimensions to accommodate high dynamics.  

We provide more architecture details for our video reconstruction approach on Bunny and UVG in Tab.~\ref{tab:archi}. $l_s\times h_s\times w_s\times dim_s$ and ~$l_d\times h_d\times w_d\times dim_d$ represent the dimensions of the static and dynamic codes, respectively. $c_1$ is the number of channels of the fused code. $Ch_{min}$ is the lowest channel width in the NeRV blocks. We adopt the settings from HNeRV~\cite{hnerv} to set the stride list, kernel size and the channel reduction rate in NeRV blocks. To match the spatial dimensions of the static codes with those of the dynamic codes, the first NeRV block performs upsampling with a upscale factor of 5. The $Conv_q$, $Conv_k$, $ Conv_v$ in CCA are all set to 2D convolution with a step size of 1, kernel size of 1, and with the number of input and output channels both set to $c_1$. 
\begin{table}[!ht]
    \centering
    \tabcolsep=0.12cm
    \resizebox{1.0\linewidth}{!}{
        \begin{tabular}{c|ccccccc}
        \hline
            Video & size & resolution & $l_s\times h_s\times w_s\times dim_s$ & $l_d\times h_d\times w_d\times dim_d$ & $c_1$ & $Ch_{min}$ & strides  \\ \hline
            Bunny & 0.35 & $640\times 1280$ & $13\times 4\times 8\times 64$ & $66\times 20\times 40\times 1$ & 36 & 16 & (5,2,2,2,2,2)  \\ 
            Bunny & 0.75 & $640\times 1280$ & $13\times 4\times 8\times 64$ & $66\times 20\times 40\times 1$ & 48 & 28 & (5,4,2,2,2)  \\ 
            Bunny & 1.5 & $640\times 1280$ & $13\times 4\times 8\times 64$ & $66\times 20\times 40\times 2$ & 70 & 38 & (5,4,2,2,2)  \\ 
            Bunny & 3 & $640\times 1280$ & $13\times 4\times 8\times 64$ & $66\times 20\times 40\times 4$ & 92 & 70 & (5,4,2,2,2)  \\ \hline
            Beauty & 3 & $960\times 1920$ & $61\times 4\times 8\times 64$ & $300\times \times20\times 40\times 2$ & 80 & 56 & (5,4,3,2,2)  \\ 
            Bosph & 3 & $960\times 1920$ & $61\times 4\times 8\times 64$ & $300\times \times20\times 40\times 2$ & 80 & 56 & (5,4,3,2,2)  \\ 
            Honey & 3 & $960\times 1920$ & $61\times 4\times 8\times 64$ & $300\times \times20\times 40\times 2$ & 80 & 56 & (5,4,3,2,2)  \\ 
            Yacht & 3 & $960\times 1920$ & $61\times 4\times 8\times 64$ & $300\times \times20\times 40\times 2$ & 80 & 56 & (5,4,3,2,2)  \\ 
            Ready & 3 & $960\times 1920$ & $31\times 4\times 8\times 64$ & $300\times \times20\times 40\times 4$ & 76 & 44 & (5,4,3,2,2)  \\ 
            Jockey & 3 & $960\times 1920$ & $31\times 4\times 8\times 64$ & $400\times \times20\times 40\times 4$ & 70 & 38 & (5,4,3,2,2)  \\ 
            Shake & 3 & $960\times 1920$ & $101\times 4\times 8\times 64$ & $150\times \times20\times 40\times 2$ & 82 & 58 & (5,4,3,2,2)  \\ \hline
        \end{tabular}
    }
    \caption{Architecture details of DS-NeRV on various tasks.}
    \label{tab:archi}
\end{table}
\section{Additional Ablation Results}
\subsection{Fusion Mechanism}
\begin{table}[!ht]
    \centering
    \tabcolsep=0.05cm
    \begin{tabular}{c|ccccccc}
    \hline
        ~ & Beauty  & Bosph & Honey & Jockey & Ready & Shake & Yacht \\ \hline
        Sum & 33.89 & 34.95 & 39.39 & 32.77 & 26.97 & 34.88 & 29.29  \\ 
        S-A & 33.80 & 34.75 & 39.45 & 31.42 & 25.69 & 35.00 & 28.85 \\ \hline
        Ours & \textbf{33.97} &\textbf{ 35.22} & \textbf{39.56} & \textbf{32.86} & \textbf{27.10} & \textbf{35.04} & \textbf{29.4}  \\ \hline
    \end{tabular}
    \caption{Ablation study for fusion mechanisms.}
    \label{tab:abla-fusion}
\end{table}
We explore different fusion mechanisms for integrating static and dynamic codes on UVG, which can be categorized into three main approaches: \textbf{a)} Summation. \textbf{b)} Spatial attention. \textbf{c)} Channel attention. As indicated in \cref{tab:abla-fusion}, a simple summation of static and dynamic codes leads to poor performance, and performing cross-spatial attention even perform worse. Cross-channel attention helps identify the most relevant channels when fusing static code and dynamic code, thereby improving the performance. 

\subsection{Spatial Dimension of the Dynamic Code}
We maintain a constant overall model size of 3M while testing the impact of varying the dimensions of dynamic codes on the results. The results, as shown in the Tab.~\ref{tab:abla-dyna_dim}, indicate that when the spatial dimension of dynamic codes is small, it becomes challenging to recover high-frequency motion information from low spatial resolution codes. Our experiments suggest that setting the dynamic codes spatial size $h_d\times w_d$ to $20\times40$ can achieve the best results.
\begin{table}[!ht]
    \centering
    \begin{tabular}{c|cc}
    \hline
        Dynamic codes dimension & PSNR & MS-SSIM \\ \hline
        $4\times8\times32$ & 37.96 & 0.9877 \\ \hline
        $4\times8\times64$ & 37.69 & 0.9869 \\ \hline
        $20\times40\times2$ & 38.55 & 0.9895 \\ \hline
        $20\times40\times4$ & \textbf{38.65} & \textbf{0.9897} \\ \hline
    \end{tabular}
    \caption{Ablation study for the dimension of dynamic codes.}
    \label{tab:abla-dyna_dim}
\end{table}

\section{Additional Quantitative Results}
\subsection{Video Reconstruction}
\begin{table}[!ht]
    \centering
    \tabcolsep=0.02cm
    \begin{tabular}{c|cccccccc}
    \hline
        ~ & Beauty  & Bosph & Honey & Jockey & Ready & Shake & Yacht  \\ \hline
        NeRV & 32.79 & 31.98 & 37.91 & 30.04 & 23.48 & 32.89 & 26.26   \\ 
        DNeRV & 31.62 & 30.18 & 33.53 & 29.62 & 22.68 & 32.45 & 25.75   \\ 
        HNeRV & 31.37 & 31.37 & 38.2 & 31.35 & 24.54 & 33.29 & 27.64   \\ \hline
        Ours & \textbf{33.29} & \textbf{34.31} & \textbf{38.98} & \textbf{32.64} & \textbf{26.41} & \textbf{34.04} & \textbf{28.72} \\ \hline
    \end{tabular}
    \caption{Video Reconstruction on UVG with 1080p}
    \label{tab:1080p_reconstruction}
\end{table}
In prior works, such as HNeRV, the resolution is adjusted to $960 \times 1920 $ to maintain a 1:2 aspect ratio, ensuring a small initial image embeddings ( $2 \times 4 $ spatial size) to improve model performance. We also conduct experiments on the standard UVG with 1080p, as shown in the~\cref{tab:1080p_reconstruction}. Our method still achieves best performance, while others suffer performance degradation due to large initial image embeddings caused by inappropriate scaling.
\subsection{Video Decoding}
\begin{table}[!ht]
    \centering
    
    \begin{tabular}{c|cccc}
    \hline
        Methods & H.264 & HEVC & NeRV  & Ours \\ \hline
        FPS & 15 & 14 & 60.08  &\textbf{ 63.54} \\ \hline
    \end{tabular}
    \caption{Decoding FPS results}
    \label{tab:decodingtime}
\end{table}
In real-world applications, inference time is a critical metric. A video is typically encoded once and requires decoding numerous times, akin to a movie that is encoded only once but viewed millions of times. Consequently, decoding time holds significance as a performance metric.
Our DS-NeRV has the advantages of faster decoding speed and does not require to do frame decoding in a sequential manner like H.264 and HEVC, as shown in Tab.~\ref{tab:decodingtime}.
\subsection{Video Inpainting}
\begin{table}[!ht]
    \centering
    \begin{tabular}{c|cccc}
    \hline
        Method & PSNR($\uparrow$) & MS-SSIM($\uparrow$) & LPIPS($\downarrow$) & FPS($\uparrow$)  \\ \hline
        IIVI & \textbf{27.66} &\textbf{ 0.9574} & 0.044 & 3.53   \\ 
        Ours & 26.45 & 0.9515 & \textbf{0.037} & \textbf{63.54 }  \\ \hline
    \end{tabular}
    \caption{Additional video inpainting results.}
    \label{tab:add_inpainting}
\end{table}
We also conduct comparative experiments on video inpainting with SOTA inpainting method IIVI~\cite{IIVI}. As shown in in \cref{tab:add_inpainting}, even without the specific design and complicated pipeline, we achieve competitive performance to contemporary SOTA inpainting methods, while achieving the fastest inference speed. IIVI requires 20 times the inpainting time we need.
\section{Additional Qualitative Results}
\subsection{Visualization of video reconstruction.}
\begin{figure*}
    \centering
    \includegraphics[width=1\linewidth]{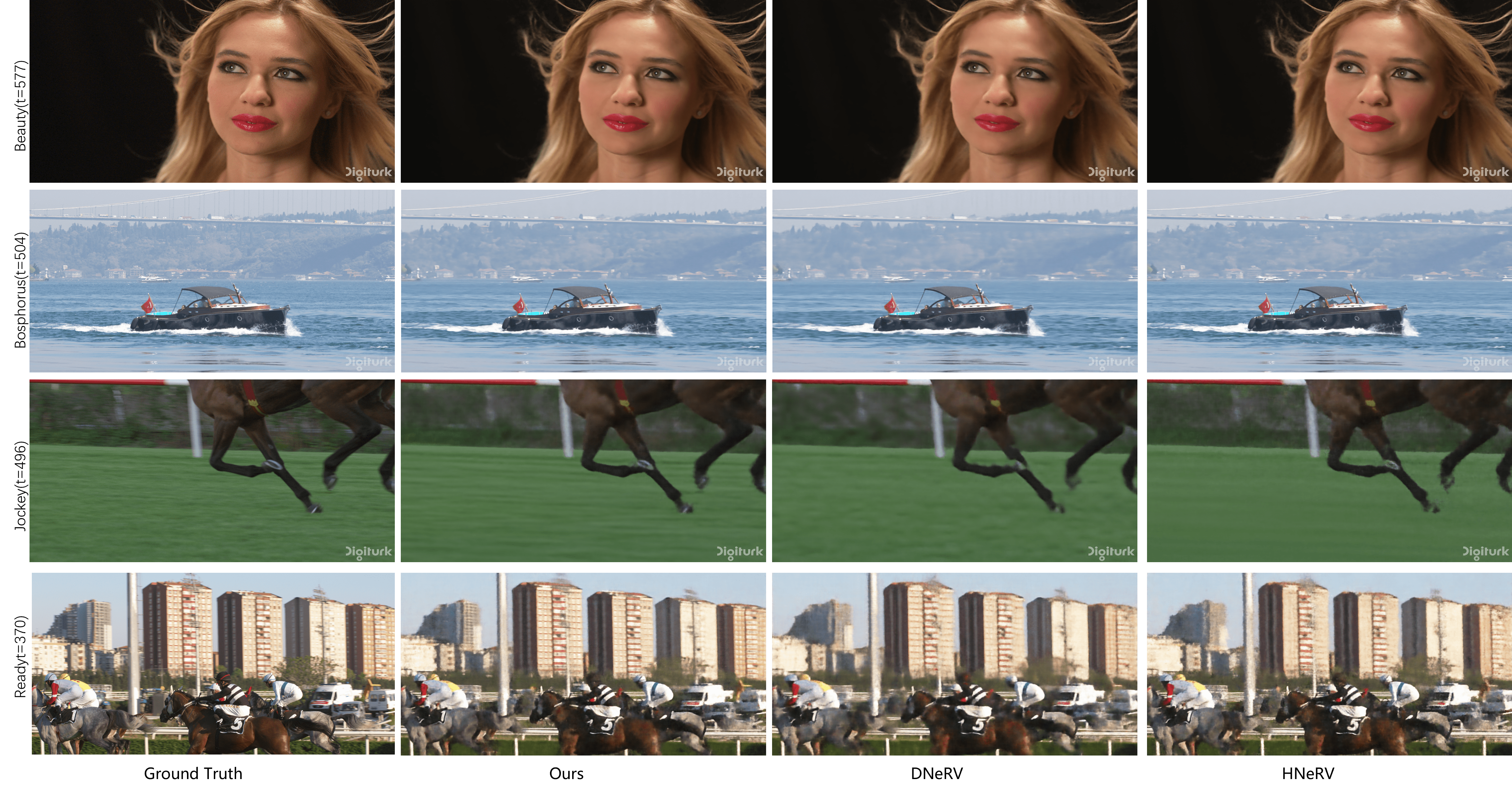}
    \caption{Additional \textbf{reconstruction} results on UVG.}
    \label{fig:addition_uvg}
\end{figure*}

\begin{figure*}
    \centering
    \includegraphics[width=1\linewidth]{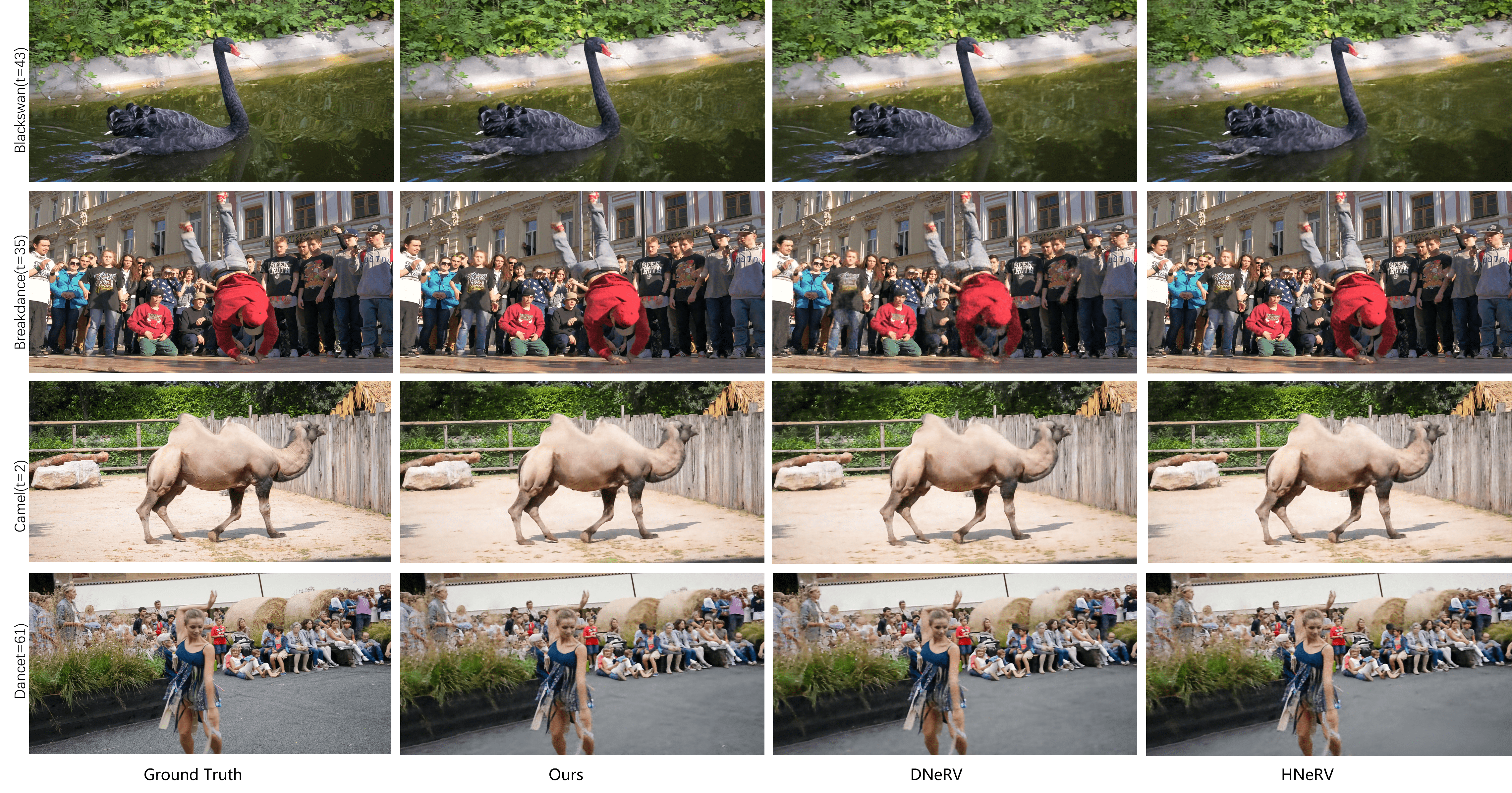}
    \caption{Additional \textbf{reconstruction} results on DAVIS.}
    \label{fig:addition_DAVIS}
\end{figure*}

We present an additional qualitative comparison of video reconstruction on UVG in Fig.~\ref{fig:addition_uvg}. DS-NeRV preserves more high frequency details compared to other methods, such as the gloss on the lips in Beauty, the distant trees on the mountains in Bosphorus, the horseshoes in Jockey, and the buildings in Ready. 
Qualitative comparisons of video reconstruction in DAVIS are also shown in Fig.~\ref{fig:addition_DAVIS}. DS-NeRV excels in reconstructing the feathers of the swan and the grassy bank in Blackswan. In Breakdance, DS-NeRV exhibits fewer artifacts in high dynamic areas, such as the dancer's shoes. DS-NeRV also provides improved reconstruction of background foliage in Camel and offers higher-quality reconstruction of the audience and grass in Dance.
\subsection{Visualization of video inpainting.}
\begin{figure*}
    \centering
    \includegraphics[width=1\linewidth]{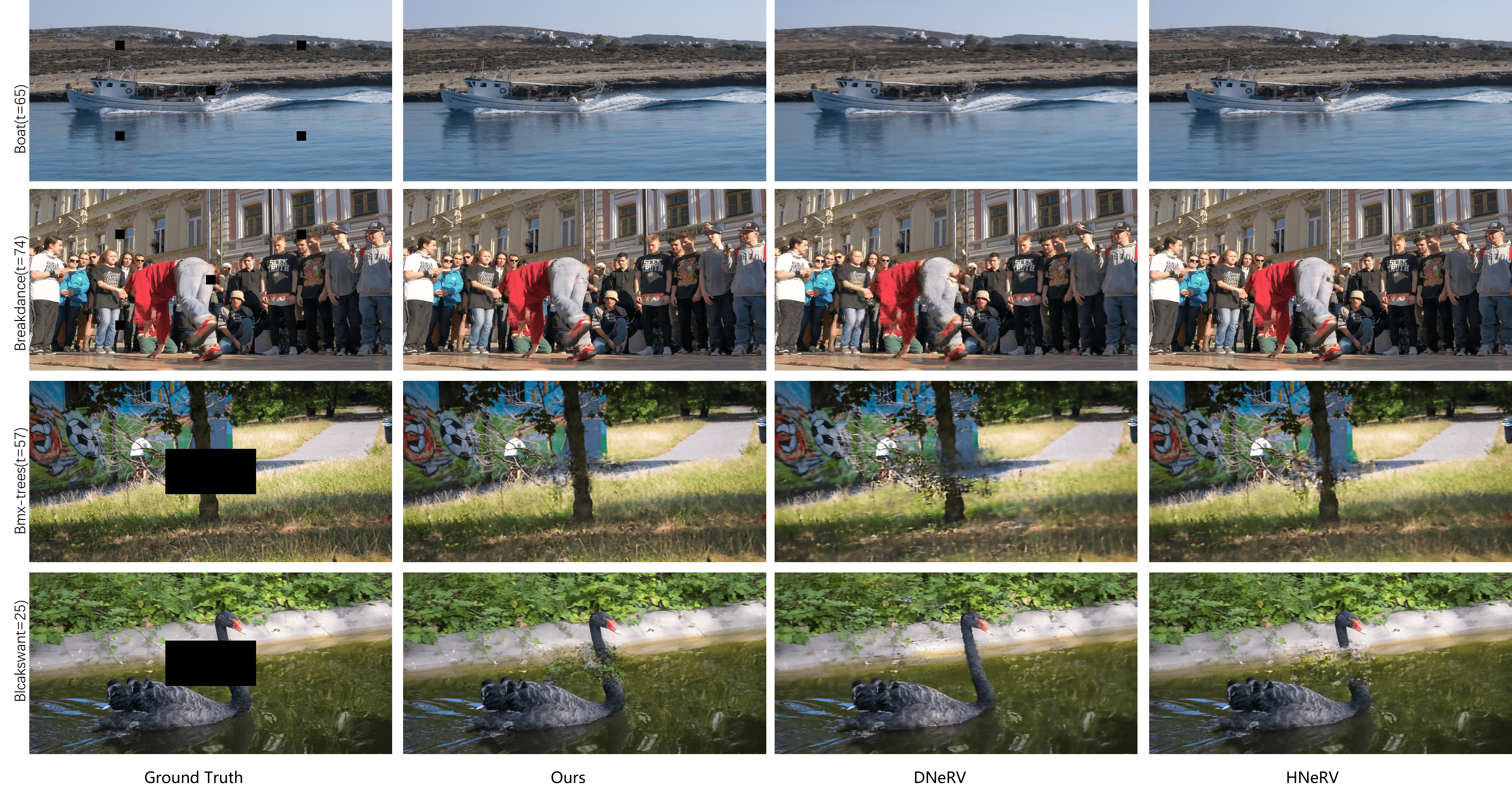}
    \caption{Additional \textbf{inpainting} results.}
    \label{fig:addition_inpaiting}
\end{figure*}

In video inpainting task, DS-NeRV is still capable of partially inferring reasonable content in the masked regions, achieving better results. Additionally, it outperforms other methods in preserving high frequency details in the rest of the image, as shown in the Fig.~\ref{fig:addition_inpaiting}.
\subsection{Visualization of video interpolation.}
\begin{figure*}
    \centering
    \includegraphics[width=1\linewidth]{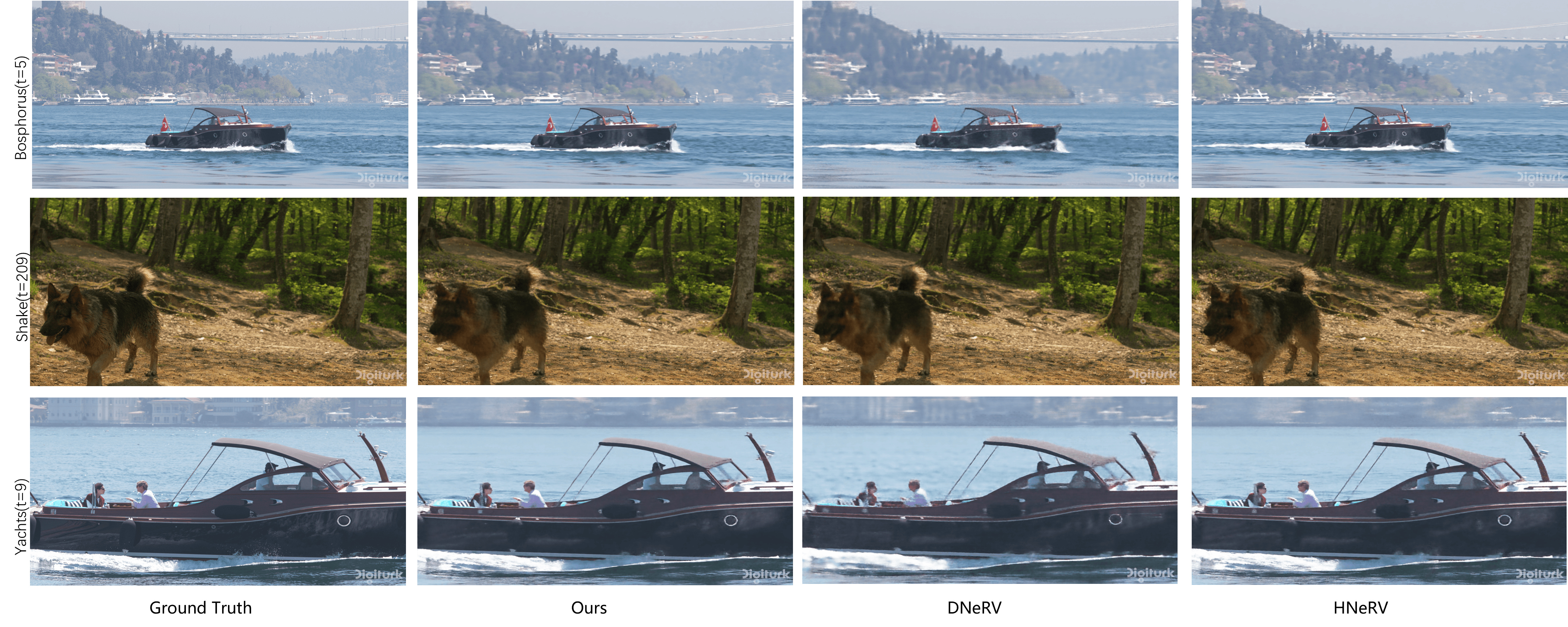}
    \caption{Additional \textbf{interpolation} results on UVG.}
    \label{fig:addition_interp}
\end{figure*}

Qualitative results for video interpolation can be viewed in Fig.~\ref{fig:addition_interp}, where all the video frames shown can not seen during training. While HNeRV and DNeRV use the frames to be interpolated itself as input to obtain embeddings during testing, our method interpolates the static and dynamic codes, obtaining the static and dynamic information of the interpolated frames, and then proceeds with decoding. DS-NeRV still achieves competitive results.

\section{Limitations and Future Work}
While DS-NeRV has achieved excellent performance, specifying the length of static and dynamic codes for each video is still a manual process during training. This provides some flexibility but limits the model's ability for adaptive adjustments. Finding the optimal static and dynamic code dimensions for each video requires time for testing. In the future, we may explore more automated and adaptive training methods to assist the model in finding the optimal static and dynamic code decomposition solutions.
\newpage

\end{document}